%% file: arxiv.tex
\documentclass{article} 
\usepackage{iclr2027_conference,times}

\input{math_commands.tex}

\usepackage{hyperref}
\usepackage{url}

\usepackage{booktabs}       
\usepackage{amsfonts}       
\usepackage{nicefrac}       
\usepackage{microtype}      
\usepackage{xcolor}         

\usepackage{pifont}
\usepackage{enumitem}

\usepackage{amsmath,amsfonts,bm}
\usepackage{multirow}
\usepackage{graphicx}
\usepackage{subcaption}
\usepackage{longtable}
\usepackage{xspace}
\usepackage[noabbrev,capitalize]{cleveref}

\newcommand{\eg}{\textit{e.g.},\xspace}
\newcommand{\pp}{\text{pp}\xspace}

\newcommand{\nmodels}{19\xspace}
\newcommand{\nfamilies}{4\xspace}
\newcommand{\ndatasets}{9\xspace}
\newcommand{\ncases}{375\xspace}
\newcommand{\ncrossfamily}{13\xspace}

\renewcommand{\paragraph}[1]{\noindent \textbf{#1}}

\title{On the Limits of Scale: \\ Inverse Scaling in Vision-Language Models}

\author{Rohit Saxena$^{1}$ \qquad Alessandro Suglia$^{1}$ \qquad Marek Rei$^{2}$ \qquad Pasquale Minervini$^{1,3}$\\
 $^{1}$School of Informatics, University of Edinburgh, United Kingdom \\  $^{2}$Imperial College London, United Kingdom \\$^{3}$Miniml.AI, United Kingdom\\
\texttt{rohit.saxena@ed.ac.uk} \\
}

\iclrfinalcopy 
\begin{document}

\maketitle

\begin{abstract}
Scaling laws suggest that increasing model size, training data, and compute generally yields monotonic performance improvements.
While this often holds at the aggregate level, we present systematic evidence that vision-language models (VLMs) exhibit \emph{inverse scaling} at the subtask level: specific question categories on which larger models perform worse than smaller counterparts within the same model family.
We evaluate \nmodels models from four distinct model families across eight multimodal benchmarks (nine evaluation splits), answering 29{,}386 questions per model, 558{,}334 model--question evaluations in total.
Using per-question pairwise analysis across all model pairs in each family, we identify \ncases subtask-model-pair combinations exhibiting inverse scaling, spanning 4{,}342 unique queries where a smaller model answers correctly but a larger model fails, and we find that these patterns replicate across independently developed architectures.
We identify \ncrossfamily subtask--dataset pairs that show inverse scaling across multiple families, consistent with the phenomenon being task-driven rather than model-specific.
The most affected domains in our study include multi-image spatial reasoning, visual perception, and university-level STEM knowledge, with accuracy gaps exceeding 20 percentage points in several cases, while diagram understanding appears comparatively resistant.
These findings challenge the assumption that increasing model size uniformly improves multimodal capabilities and suggest that model selection 
should be task-aware\footnote{Our code is available at \href{https://github.com/saxenarohit/vlm_inverse_scaling}{https://github.com/saxenarohit/vlm\_inverse\_scaling}}.
\end{abstract}
%
\section{Introduction} \label{sec:intro}
\begin{figure*}[t]
\centering
\includegraphics[width=\textwidth]{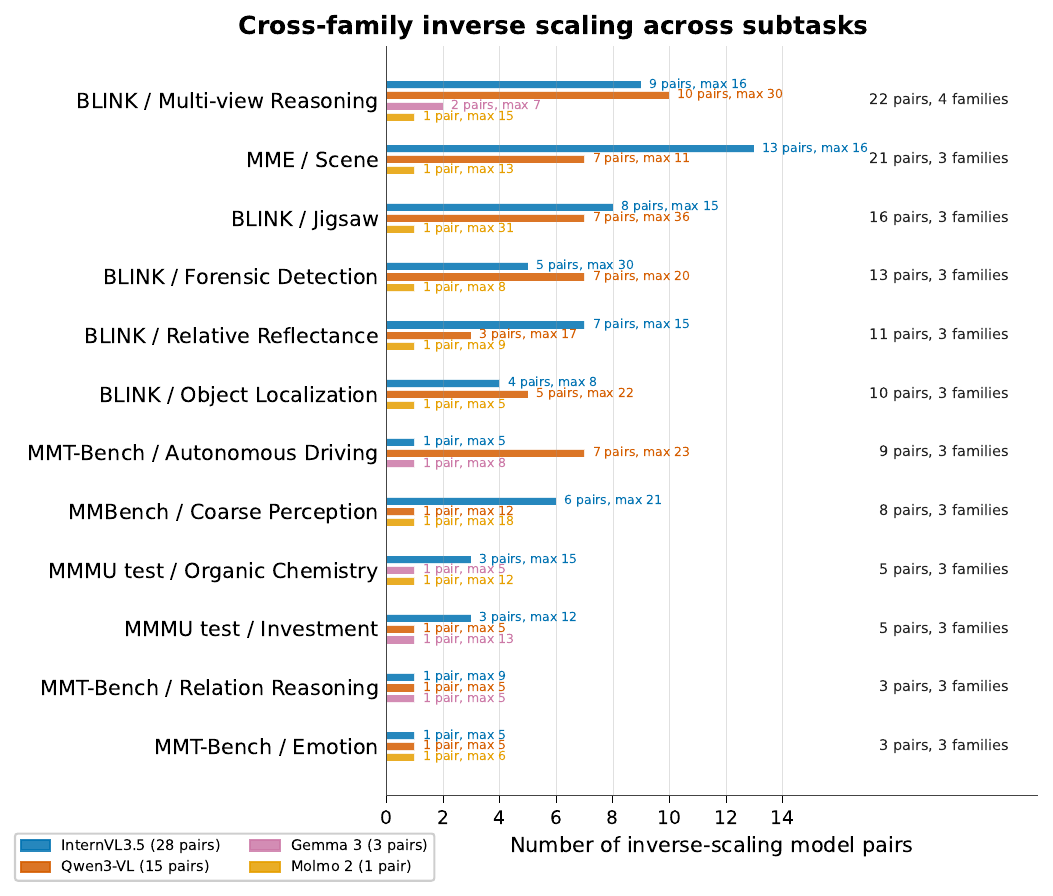}
\caption{Cross-family inverse scaling across subtasks. Each bar shows the number of pairwise inverse-scaling cases for a given model family on that subtask; the annotation shows the count and the strongest case's Net flip. Right column: total cases and number of families. Only the 12 subtasks with the most cases are shown; all 13 appear in \cref{tab:cross_family}.}
\label{fig:cross_family}
\end{figure*}

Neural scaling laws suggest that model performance generally improves with increasing model size, data, and compute \citep{kaplan2020scaling, hoffmann2022training}.
This empirical trend has guided the development of foundation models and reinforced a simple working assumption: \textit{larger models should perform better}.
However, recent work shows that this assumption can fail.
In language models, \citet{mckenzie2023inverse} identify \emph{inverse scaling} tasks on which performance worsens as model scale increases, while \citet{wei2023ushaped} show that some of these trends are better described as non-monotonic or U-shaped behaviour rather than uniformly decreasing.
More recently, \citet{gema2025testtimecompute} demonstrate an analogous phenomenon for reasoning models, where increasing test-time compute can also reduce accuracy.
In multimodal reasoning, concurrent work by \citet{bae2025reasoningvisionlanguagemodelsinversely} shows that adding visual distractors can reduce VLMs' accuracy and induce inverse-scaling-like behaviour, though in a constructed distractor setting rather than in standard benchmark evaluation.
Whether inverse scaling also arises in VLMs under standard benchmark evaluation, without artificially introducing distractors, remains an open question.
VLMs inherit the same scaling paradigm as language models, with model families typically released in sizes ranging from small to very large.
At the same time, VLMs must jointly process visual and textual information, creating additional opportunities for failure~\citep{asadi2026mirage, saxena2026vlm}.
A larger model may achieve higher overall benchmark accuracy while becoming less accurate on particular forms of visual reasoning.
If so, standard aggregate evaluation may obscure regressions that are scientifically important and practically relevant for deployment.
Such regressions are already visible, though unremarked, in providers' own reports: in the Qwen3-VL technical report~\citep{qwen3vl}, the 8B instruct model scores above both the 32B and the 30B-A3B on BLINK (69.1 vs.\ 67.3 and 67.7; Tables 3--4), the 32B scores above the 235B-A22B on VLMs-Are-Blind (87.0 vs.\ 80.4; Tables 2--3), and CountBench accuracy declines monotonically from 2B to 4B to 8B (88.4, 84.9, 80.5; Table 4). Aggregate reporting records these numbers without identifying which capabilities regress or on which questions.
In this work, we conduct a systematic study of subtask-level inverse scaling across model sizes in vision-language models on standard multimodal benchmarks, where subtasks are the category labels provided by each benchmark (e.g., ``Jigsaw'' in BLINK~\citep{fu2024blink} or ``Organic Chemistry'' in MMMU~\citep{yue2023mmmu}; see \cref{tab:setup} for the full list of benchmarks).
Our analysis is structured around \emph{within-family} comparisons: we compare models released under the same family name, which generally share a high-level architecture and training recipe but may differ across sizes in language backbone, vision encoder, active parameter count, training data, alignment procedure, or inference-time visual processing.
This design reduces, but does not eliminate, confounds relative to comparisons across independently developed model families.
We evaluate \nmodels models from four distinct model families (Qwen3-VL~\citep{qwen3vl}, InternVL3.5~\citep{internvl35}, Gemma~3~\citep{gemma3}, and Molmo~2~\citep{molmo2}) spanning roughly 1B to 235B total parameters and including both dense and Mixture-of-Experts (MoE) variants.
Across eight established multimodal benchmarks used for evaluating frontier VLMs (nine evaluation splits), this yields 29{,}386 questions per model.
Rather than relying solely on aggregate accuracy, we introduce a \emph{per-question flip analysis} that compares the behaviour of paired models on the same evaluation items, measuring how often a smaller model answers correctly when a larger model fails, and vice versa.
This paired analysis quantifies inverse scaling at the subtask level, revealing patterns that are often hidden in benchmark-level summaries.
\begin{figure*}[tb]
\centering
\includegraphics[width=\textwidth]{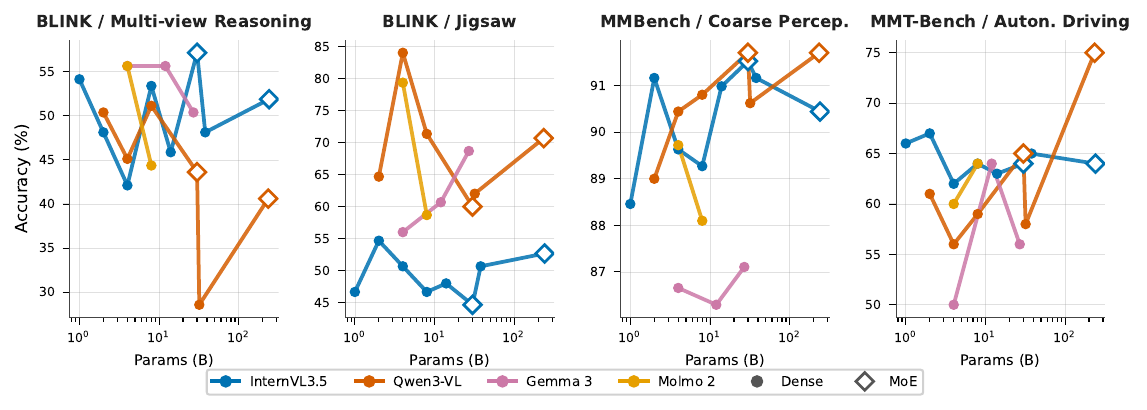}
\caption{Scaling curves for four cross-family subtasks. Filled circles: dense models; open diamonds: MoE variants. All show non-monotonic behaviour across families.}
\label{fig:scaling_curves}
\end{figure*}
Our results show that inverse scaling is widespread at the subtask level (\cref{fig:scaling_curves}).
Across all pairwise comparisons within each family, we identify \ncases subtask--model-pair cases exhibiting inverse scaling, spanning all four model families and appearing in every benchmark we evaluate.
These patterns are not confined to a single architecture or training recipe: \ncrossfamily subtask--dataset pairs exhibit inverse scaling across multiple families (\cref{fig:cross_family}), consistent with a task-driven rather than purely model-specific phenomenon.
The strongest effects arise in multi-image spatial reasoning (BLINK), multi-task visual understanding (MMT-Bench), and university-level STEM problem solving (MMMU), whereas diagram understanding (AI2D) appears comparatively resistant.
Crucially, these findings persist when restricting to dense-only model pairs (193 cases, 8 cross-family subtasks), confirming that the phenomenon is not driven solely by MoE models.
\cref{app:examples} presents qualitative examples illustrating the range of failure modes, from misidentifying clearly visible objects to selecting plausible but incorrect options.
These findings challenge the assumption that increasing scale uniformly improves multimodal capability.
More broadly, they suggest that model choice should depend not only on aggregate benchmark performance, but also on the specific capabilities required by the target application domain~\citep{liu2025towards}.
%

\section{Related Work} \label{sec:related}

\paragraph{Scaling laws.}
Scaling-law work shows that language-model loss and many downstream capabilities improve predictably as parameters, data, and compute increase~\citep{kaplan2020scaling,hoffmann2022training}.
Similar empirical regularities have also been reported in vision-language pretraining: \citet{cherti2023reproducible} find power-law behaviour for CLIP-style contrastive models across zero-shot classification, retrieval, linear probing, and fine-tuning.
Similarly, \citet{aghajanyan2023scaling} reports relevant scaling laws for multimodal models.
While these works focus on aggregate task performance, we analyse whether monotonic improvements persist at the finer-grained level of benchmark subtasks within a VLM family, which we assume correlates more directly with specific multimodal skills.

\paragraph{Inverse scaling in LLMs and VLMs.}
The \emph{Inverse Scaling Prize}~\citep{mckenzie2023inverse} introduced curated text-only tasks on which larger language models performed worse and identified causes such as strong priors, unwanted imitation, and distractor tasks.
\citet{wei2023ushaped} expanded the model range and found that many of these curves become U-shaped rather than monotonically decreasing, and that one-shot examples or chain-of-thought prompting can mitigate some failures.
Along a different axis, \citet{gema2025testtimecompute} construct tasks on which extending the reasoning length of large reasoning models degrades accuracy, yielding inverse scaling in test-time compute rather than model size.
Concurrently, \citet{bae2025reasoningvisionlanguagemodelsinversely} study how visual and textual distractors affect multimodal reasoning, finding that visual distractors reduce VLM accuracy without lengthening reasoning chains, a mechanism distinct from text-only distractor effects.

Our work complements these efforts by examining open VLM families on standard multimodal benchmarks rather than on specially constructed tasks and by detecting inverse scaling via per-question pairwise flips within existing subtasks.
In line with \cite{mckenzie2023inverse}, our findings are observational evidence from publicly available VLM checkpoints.
%
%

\section{Methodology: Investigating Inverse Scaling in VLMs} \label{sec:method}
\paragraph{Problem Formulation.}
Let $\mathcal{F} = \{f_1, f_2, \ldots, f_K\}$ denote a family of VLMs ordered by size, measured in terms of the number of parameters, where $f_1$ is the smallest and $f_K$ the largest.~\footnote{
We order models by total parameters. Because MoE and dense variants differ in active parameters, \cref{tab:setup} reports both counts; \cref{tab:dense_cases} repeats the analysis on dense-only pairs and \cref{app:size_ratio} on pairs with at least twice the active parameters.
}
Let $\mathcal{D}$ be a question-answering (QA) benchmark partitioned into subtasks $\{S_1, S_2, \ldots, S_T\}$, where each subtask is a set of evaluation questions of the same type (\eg a subject area or reasoning category), and let $\text{acc}(f_k, S_t)$ denote the accuracy of model $f_k$ on subtask $S_t$.
Under a monotonic scaling assumption, one would expect $\text{acc}(f_k, S_t)$ to be non-decreasing in $k$.
We say subtask $S_t$ exhibits \emph{inverse scaling} for the pair $(f_i, f_j)$ with $i < j$ iff $\text{acc}(f_i, S_t) > \text{acc}(f_j, S_t)$.
\paragraph{Per-Question Flip Analysis.}
To quantify inverse scaling at a finer granularity than subtask-level accuracy, we compare individual question outcomes across model pairs.
For a pair of models $(f_i, f_j)$ with $i < j$ and a subtask $S_t$, let $Q = \{q_1, \ldots, q_n\} \subseteq S_t$ denote the questions in that subtask.
For each question $q \in Q$, let $c_i(q), c_j(q) \in \{0, 1\}$ be correctness indicators, where 1 means correct, and 0 means incorrect.
We then define:
\begin{align}
    \text{FlipInv} &= \sum_{q \in Q} \mathbf{1}[c_i(q) = 1 \wedge c_j(q) = 0] && \text{(smaller correct, larger wrong)} \label{eq:flipinv} \\
    \text{FlipNorm} &= \sum_{q \in Q} \mathbf{1}[c_i(q) = 0 \wedge c_j(q) = 1] && \text{(larger correct, smaller wrong)} \label{eq:flipnorm} \\
    \text{Net} &= \text{FlipInv} - \text{FlipNorm} && \text{(net advantage for the smaller model)} \label{eq:net}
\end{align}
\noindent Here, FlipInv counts disagreements that favor the smaller model $f_i$, FlipNorm counts disagreements that favor the larger model $f_j$, and Net is the difference between the two.
A positive Net means that, among discordant questions, the smaller model is correct more often than the larger one.
We apply this analysis to \emph{all} within-family model pairs.
A family with $K$ models therefore contributes $\binom{K}{2}$ pairwise comparisons for each subtask.
%
\section{Experimental Setup}
\label{sec:experiments}
\paragraph{Models.}
We evaluate \nmodels models across \nfamilies open-weight families~\footnote{We exclude closed-weight models, whose parameter counts and configurations are unknown.}: Qwen3-VL~\citep{qwen3vl}, InternVL3.5~\citep{internvl35}, Gemma~3~\citep{gemma3}, and Molmo~2~\citep{molmo2}, summarized in \cref{tab:setup}~(left). Each family provides a range of model sizes, including both dense architectures and Mixture-of-Experts (MoE) variants. The families were selected to maximize architectural diversity with at least two sizes per family.
\begin{table*}[t]
\caption{\textbf{Left:} Model families evaluated; $^*$MoE models with active parameters in parentheses. \textbf{Right:} Evaluation benchmarks; subtask counts reflect the category granularity used in our analysis. $^\dagger$ScienceQA is filtered to the 2{,}017 questions that include an image, excluding text-only questions.}
\label{tab:setup}
\centering
\small
\begin{minipage}[t]{0.45\textwidth}
\centering
\setlength{\tabcolsep}{3pt}
\begin{tabular}{ll}
\toprule
\textbf{Family} & \textbf{Models} \\
\midrule
Qwen3-VL & 2B, 4B, 8B, 30B-A3B$^*$\!{\scriptsize(3B)}, \\
         & 32B, 235B-A22B$^*$\!{\scriptsize(22B)} \\
\midrule
InternVL3.5 & 1B, 2B, 4B, 8B, 14B, \\
            & 30B-A3B$^*$\!{\scriptsize(3B)}, 38B, \\
            & 241B-A28B$^*$\!{\scriptsize(28B)} \\
\midrule
Gemma~3 & 4B, 12B, 27B \\
\midrule
Molmo~2 & 4B, 8B \\
\bottomrule
\end{tabular}
\end{minipage}
\hfill
\begin{minipage}[t]{0.52\textwidth}
\centering
\setlength{\tabcolsep}{3pt}
\begin{tabular}{llccc}
\toprule
\textbf{Dataset} & \textbf{Split} & \textbf{N} & \textbf{Sub.} & \textbf{Type} \\
\midrule
MMT-Bench & VAL & 3,127 & 32 & MCQ \\
MMMU & dev\_val & 1,050 & 132 & MCQ+open \\
MMMU & test & 10,500 & 177 & MCQ+open \\
ScienceQA$^\dagger$ & test & 2,017 & 62 & MCQ \\
MME & test & 2,374 & 14 & Yes/No \\
AI2D & test & 3,088 & 15 & MCQ \\
MathVista & testmini & 1,000 & 5 & MCQ+free \\
BLINK & val & 1,901 & 14 & MCQ \\
MMBench & DEV\_EN & 4,329 & 20 & MCQ \\
\bottomrule
\end{tabular}
\end{minipage}
\end{table*}

\paragraph{Datasets.}
We evaluate on \ndatasets benchmark splits covering diverse multimodal capabilities, summarized in \cref{tab:setup}~(right). The suite covers perception and cognition (MME~\citep{fu2023mme}), diagram understanding (AI2D~\citep{kembhavi2016diagram}), K--12 science (ScienceQA~\citep{lu2022scienceqa}), multi-image visual perception (BLINK~\citep{fu2024blink}), general multimodal reasoning (MMT-Bench~\citep{ying2024mmtbench} and MMBench~\citep{liu2023mmbench}), university-level knowledge (MMMU~\citep{yue2023mmmu}), and mathematical reasoning (MathVista~\citep{lu2023mathvista}). Because these benchmarks are widely used to compare released VLMs, subtask-level regressions are practically relevant to model selection.
In total, we perform 171 inference runs (\nmodels models $\times$ \ndatasets evaluation splits), evaluating 29{,}386 unique questions per model (558{,}334 model--question evaluations). Full model identifiers, compute details, and API configurations are provided in \cref{app:details}.

\subsection{Evaluation Protocol}
\label{sec:protocol}
\paragraph{Inference.} All models are evaluated zero-shot with direct prompting and greedy decoding, with thinking modes disabled to separate model scale from test-time compute~\citep{gema2025testtimecompute}; MCQ prompts ask for the option letter and open-ended prompts for a concise answer (settings in \cref{app:generation}, prompts in \cref{app:prompts}). Answers are verified as in VLMEvalKit~\citep{duan2025vlmevalkitopensourcetoolkitevaluating}: deterministic letter extraction and exact or numeric matching, with GPT-5.4 as a multimodal fallback adjudicator when these rules are inconclusive (0.4\% of responses).

\paragraph{Case selection.} We report a pairwise inverse-scaling case when the subtask contains at least $n \geq 20$ shared questions and the net flip advantage is $\text{Net} \geq 5$. The first threshold excludes subtasks too small for meaningful comparison; the second excludes cases where the asymmetry is trivially small. The findings are robust to stricter thresholds: 132 cases across 44 subtask--dataset pairs remain at Net $\geq 10$, with 4 cross-family subtasks still replicated (\cref{app:net_dist}).

\section{Results}
\label{sec:results}

\subsection{Aggregate Scaling Behavior}

At the aggregate level, the largest model in each family achieves the highest or near-highest accuracy on most benchmarks (full results in \cref{app:aggregate}). However, several clear exceptions stand out. On BLINK, Qwen 30B-A3B (62.4\%) substantially underperforms smaller dense models such as 4B (66.4\%), and Molmo 4B (63.5\%) outperforms 8B (59.3\%). Molmo 4B also outperforms 8B on MMBench (88.3\% vs.\ 87.9\%).
\subsection{Subtask-Level Inverse Scaling}

While aggregate performance scales mostly monotonically, the subtask level tells a different story. Inverse scaling appears across all four families and every benchmark we evaluate.

\begin{table}[t]
\caption{ Pairwise inverse scaling cases by family and dataset. Each cell reports the number of detected cases over all eligible model-pair--subtask comparisons ($n \geq 20$ shared questions); a case is a unique (subtask, smaller model, larger model) tuple meeting the thresholds in \cref{sec:experiments}.}
\label{tab:pairwise_summary}
\centering
\small
\setlength{\tabcolsep}{4pt}
\begin{tabular}{lrrrrr}
\toprule
\textbf{Dataset} & \textbf{Qwen3-VL} & \textbf{InternVL} & \textbf{Gemma~3} & \textbf{Molmo~2} & \textbf{Total} \\
\midrule
MMT-Bench & 31/480 & 49/896 & 6/96 & 1/32 & 87/1504 \\
MMMU (dev\_val+test) & 34/1620 & 67/3024 & 8/324 & 6/108 & 115/5076 \\
ScienceQA & 7/450 & 0/840 & 10/90 & 0/30 & 17/1410 \\
MME & 10/210 & 23/392 & 0/42 & 2/14 & 35/658 \\
AI2D & 0/195 & 8/364 & 0/39 & 1/13 & 9/611 \\
MathVista & 1/75 & 2/140 & 0/15 & 0/5 & 3/235 \\
BLINK & 37/210 & 48/392 & 2/42 & 8/14 & 95/658 \\
MMBench & 2/90 & 10/168 & 0/18 & 2/6 & 14/282 \\
\midrule
\textbf{Total} & \textbf{122/3330} & \textbf{207/6216} & \textbf{26/666} & \textbf{20/222} & \textbf{375/10434} \\
\bottomrule
\end{tabular}
\end{table}

\paragraph{Breadth across families and benchmarks.} 
%
\cref{tab:pairwise_summary} summarizes prevalence, not model quality (aggregate accuracies: \cref{app:aggregate}; strongest cases: \cref{tab:top_cases}; all cases: \cref{app:pairwise}). InternVL contributes the most cases (207), reflecting its eight models and 6{,}216 eligible comparisons; normalized, Qwen3-VL (3.7\%), InternVL (3.3\%), and Gemma~3 (3.9\%) have similar rates, and Molmo~2 a higher one (9.0\%, from a single pair). Raw counts are highest for MMMU (115), BLINK (95), and MMT-Bench (87) and lowest for AI2D (9) and MathVista (3); normalized, BLINK leads (14.4\%), followed by MMT-Bench (5.8\%), MME (5.3\%), and MMBench (5.0\%), with MMMU at 2.3\% and ScienceQA at 1.2\%. Because $\mathrm{Net} = 5$ corresponds to an accuracy gap of 0.45\pp at $n = 1{,}109$ but 25\pp at $n = 20$, we read counts and rates descriptively, not as intrinsic benchmark susceptibility. The 375 cases span 115 subtask--dataset pairs and 171 family--subtask combinations. Adjacent-size comparisons alone (e.g., 4B$\to$8B, 8B$\to$14B) yield 197 cases across 107 subtask--dataset pairs, including 12 cross-family subtasks, so the patterns include local regressions, not only distant sizes. Of the 171 family--subtask pairs, 87\% have non-monotonic accuracy curves rather than consistent degradation, mirroring U-shaped trends in language-model inverse scaling~\citep{wei2023ushaped}. The findings are qualitatively stable across Net thresholds (\cref{app:net_dist}).

\paragraph{Statistical significance.} As a descriptive measure of paired asymmetry, we report one-sided McNemar exact $p$-values for each candidate case, i.e., for each subtask--model-pair comparison, computed over the subtask's shared questions. Because these tests are performed on many overlapping model-pair/subtask combinations and because cases are selected using the same observed data, these $p$-values should be interpreted as exploratory rather than confirmatory. With this caveat, 133 of 375 cases (35.5\%) have uncorrected $p < 0.05$, rising from 10.4\% of cases with Net 5--7 to 100\% with Net $\geq 21$; \cref{app:significance} reports the Benjamini--Hochberg correction over the reported cases and over the full space of eligible comparisons from which they were selected. Cross-family replication provides complementary robustness evidence that the patterns are not confined to a single family (\cref{sec:cross_family}).

\paragraph{Strongest cases.} \cref{tab:top_cases} presents the strongest individual cases. BLINK dominates: Jigsaw appears three times in the top 10 across the Qwen and Molmo families, led by Qwen 4B outperforming 30B-A3B by 24.0\pp (Net = 36), and Forensic Detection three times for InternVL, where cases \#4, \#6, and \#7 have FlipNorm = 0: no single question where the MoE model succeeds and the smaller model fails. The only non-BLINK entries are MME Landmark (InternVL 1B vs.\ 2B) and ScienceQA Basic Economic Principles, where Gemma~3 4B outperforms 27B by 39.4\pp. Accuracy gaps range from 7.2\pp to 39.4\pp among the top 10. The six BLINK cases among them keep a positive Net under every deterministic reordering of their answer options (\cref{app:option_order}).
\begin{table}[t]
\caption{Top 10 pairwise cases ranked by Net (FlipInv $-$ FlipNorm); Acc $\Delta$ in pp.}
\label{tab:top_cases}
\centering
\small
\setlength{\tabcolsep}{3pt}
\begin{tabular}{clllccccccc}
\toprule
\textbf{\#} & \textbf{Family} & \textbf{Dataset} & \textbf{Subtask} & \textbf{Small} & \textbf{Large} & $n$ & \textbf{FI} & \textbf{FN} & \textbf{Net} & \textbf{Acc $\Delta$} \\
\midrule
1 & Qwen3-VL & BLINK & Jigsaw & 4B & 30B-A3B & 150 & 45 & 9 & 36 & 24.0 \\
2 & Qwen3-VL & BLINK & Jigsaw & 4B & 32B & 150 & 44 & 11 & 33 & 22.0 \\
3 & Molmo~2 & BLINK & Jigsaw & 4B & 8B & 150 & 43 & 12 & 31 & 20.7 \\
4 & InternVL & BLINK & Forensic Detection & 2B & 30B-A3B & 132 & 30 & 0 & 30 & 22.7 \\
5 & Qwen3-VL & BLINK & Multi-view Reasoning & 8B & 32B & 133 & 38 & 8 & 30 & 22.6 \\
6 & InternVL & BLINK & Forensic Detection & 4B & 30B-A3B & 132 & 29 & 0 & 29 & 22.0 \\
7 & InternVL & BLINK & Forensic Detection & 14B & 30B-A3B & 132 & 29 & 0 & 29 & 22.0 \\
8 & Qwen3-VL & BLINK & Multi-view Reasoning & 2B & 32B & 133 & 37 & 8 & 29 & 21.8 \\
9 & InternVL & MME & Landmark & 1B & 2B & 400 & 55 & 26 & 29 & 7.2 \\
10 & Gemma~3 & ScienceQA & Basic Economic Principles & 4B & 27B & 71 & 31 & 3 & 28 & 39.4 \\
\bottomrule
\end{tabular}
\end{table}
%
\subsection{Per-Family Analysis}

\textbf{Qwen3-VL (122 cases)}. The most striking inverse scaling appears on BLINK, where the MoE model 30B-A3B consistently underperforms dense models. On Jigsaw, the 4B model achieves 84.0\% accuracy compared to 30B-A3B's 60.0\% (+24.0\pp, Net = 36). On Multi-view Reasoning, multiple smaller models outperform the 32B model by over 20\pp.

\textbf{InternVL3.5 (207 cases)}. InternVL has the largest raw count (207), partly reflecting its larger family, while its normalized rate (3.3\%) is comparable to Qwen (3.7\%) and Gemma (3.9\%). Key transitions include 8B $\rightarrow$ 14B (where 14B degrades on MMMU subtasks, MME categories, and BLINK), and 38B $\rightarrow$ 241B-A28B (the MoE model fails to improve on many specialized subtasks). On BLINK Forensic Detection, four dense models (2B--14B) outperform 30B-A3B with FlipNorm = 0.

\textbf{Gemma~3 (26 cases)}. Despite having only 3 models (and thus $\binom{3}{2} = 3$ pairs), Gemma~3 shows 26 cases. ScienceQA contributes 10 cases, including the striking ``Particle motion and energy'' where 12B achieves 95.1\% vs.\ 27B's 51.2\% (+43.9\pp, Net = 18).

\textbf{Molmo~2 (20 cases)}. With only 2 models (4B and 8B), Molmo~2 shows 20 cases. BLINK contributes 8 cases, with Jigsaw showing Net = 31 (4B: 79.3\% vs.\ 8B: 58.7\%). The 4B model outperforms 8B at the aggregate level on both BLINK (63.5\% vs.\ 59.3\%) and MMBench (88.3\% vs.\ 87.9\%).
%
\subsection{Cross-Family Patterns}
\label{sec:cross_family}

Cross-family replication provides additional evidence that inverse scaling is systematic. 
We first identify pairwise cases using the question-level criteria in \cref{sec:experiments}. Separately, for cross-family summarization, we highlight subtask--dataset pairs that recur in at least three of the four families. This criterion is a reporting choice for emphasizing the strongest replications, not part of the underlying pairwise case definition. \cref{tab:cross_family} presents the \ncrossfamily subtask--dataset pairs meeting this criterion, organized by dataset. An additional 29 subtasks appear in exactly two families (\cref{app:cross_family_2}).

\paragraph{BLINK dominates cross-family inverse scaling.} Five of 14 BLINK subtasks show cross-family inverse scaling, with Multi-view Reasoning appearing in all four families (Qwen, InternVL, Gemma~3, Molmo~2). These subtasks involve spatial reasoning across multiple images. BLINK contributes high Max Net values (up to 36), with multi-image spatial reasoning showing the strongest inverse scaling effects. \cref{fig:scaling_curves} shows representative scaling curves for four cross-family subtasks, illustrating the non-monotonic accuracy patterns across families.

\paragraph{MMBench coarse perception.} Coarse perception shows inverse scaling in Qwen, InternVL, and Molmo~2 (Max Net = 21).

\paragraph{MMMU STEM subjects.} Organic Chemistry (Gemma~3, InternVL, Molmo~2), Investment (Gemma~3, InternVL, Qwen), and Thermodynamics (InternVL, Molmo~2, Qwen) show cross-family inverse scaling on university-level problems. These require specialized domain knowledge combined with visual understanding of diagrams and figures.

\paragraph{MMT-Bench and MME.} Autonomous Driving on MMT-Bench (3 families, Max Net = 23) and Scene on MME (3 families, Max Net = 16) also show cross-family replication.

%
\subsection{Other Findings}

\paragraph{MoE models frequently appear in inverse scaling pairs.}
Mixture-of-Experts models frequently appear as the larger model in inverse scaling pairs. Qwen 30B-A3B (3B active parameters) underperforms the dense 32B model on several BLINK subtasks. InternVL 241B-A28B (28B active) fails to improve over the dense 38B on AI2D (both 87.1\%) and degrades on several BLINK subtasks. However, MoE and dense models differ in active parameter count, training recipe, and alignment procedure, so we cannot isolate the contribution of expert routing from these confounds.

The findings above show that inverse scaling is widespread and cross-family, but is it simply the largest model underperforming, or a more complex relationship between scale and capability? Three observations suggest the latter.

\paragraph{Scaling trajectories are non-monotonic.}
Rather than a single large-model drop, many subtasks exhibit multiple peaks and valleys across the size range. For example, InternVL's performance on ``Mechanics of Materials'' on MMMU shows: 1B = 27.6\%, 2B = 48.3\%, 4B = 44.8\%, 8B = 31.0\%, 14B = 51.7\%, 30B-A3B = 58.6\%, 38B = 62.1\%, 241B-A28B = 27.6\%. This suggests that different model sizes have qualitatively different strengths rather than uniformly improving capabilities.

\paragraph{In extreme cases, the smallest model wins.}
This non-monotonicity can be so pronounced that the smallest model in a family achieves the highest subtask accuracy. Qwen 2B is the best on MMMU test AI Content Detection (65.0\% vs.\ 235B's 45.0\%). Molmo 4B outperforms 8B on all BLINK subtasks showing inverse scaling. These cases show that inverse scaling is not limited to marginal regressions at the top of the size range.

\paragraph{Inverse scaling survives a twofold increase in model size.}
Restricting to pairs in which the larger model has at least twice the total parameters of the smaller leaves 326 of the 375 cases (3.3\% of 9{,}768 comparisons). Using active parameters, which matters for the MoE models, leaves 239 cases (2.9\% of 8{,}214 comparisons). Both restrictions keep 11 of the 13 cross-family subtasks (\cref{app:size_ratio}). At the largest size gap in each family (smallest vs.\ largest model, on every eligible subtask), the smaller model is ahead by Net~$\geq 5$ in 9 of 666 comparisons (Qwen3-VL 2B vs.\ 235B-A22B: 4; InternVL 1B vs.\ 241B-A28B: 0; Gemma~3 4B vs.\ 27B: 5); Molmo~2 has a single pair. Cases therefore persist at twofold size gaps but are limited across the full size range.

\paragraph{The pattern persists without MoE models.}
MoE architectures, with fewer active parameters and distinct routing, could drive the phenomenon. We therefore repeat the pairwise analysis on dense pairs only, excluding all MoE variants (Qwen 30B-A3B and 235B-A22B; InternVL 30B-A3B and 241B-A28B). This still yields 193 pairwise cases (\cref{tab:dense_cases,tab:dense_cross}), with 8 subtask--dataset pairs replicating across at least three families, including BLINK Multi-view Reasoning, which appears in all four. Inverse scaling is therefore not solely an MoE artifact.
\begin{table}[t]
\caption{Dense-only pairwise inverse scaling cases by family and dataset (MoE models excluded). Gemma~3 and Molmo~2 contain no MoE models, so their counts equal those of the full analysis.}
\label{tab:dense_cases}
\centering
\small
\begin{tabular}{lccccc}
\toprule
\textbf{Dataset} & \textbf{Qwen3-VL} & \textbf{InternVL} & \textbf{Gemma~3} & \textbf{Molmo~2} & \textbf{Total} \\
\midrule
MMT-Bench & 10 & 20 & 6 & 1 & 37 \\
MMMU (dev\_val+test) & 13 & 40 & 8 & 6 & 67 \\
ScienceQA & 4 & 0 & 10 & 0 & 14 \\
MME & 3 & 14 & 0 & 2 & 19 \\
AI2D & 0 & 1 & 0 & 1 & 2 \\
MathVista & 0 & 1 & 0 & 0 & 1 \\
BLINK & 13 & 23 & 2 & 8 & 46 \\
MMBench & 0 & 5 & 0 & 2 & 7 \\
\midrule
\textbf{Total} & \textbf{43} & \textbf{104} & \textbf{26} & \textbf{20} & \textbf{193} \\
\bottomrule
\end{tabular}
\end{table}
\subsection{Error Taxonomy}
\label{sec:taxonomy}

To understand why larger models fail on inverse-scaling questions, we use GPT-5.4 as a multimodal annotator to assign failure categories to inverse-scaling flip examples. We sample up to 10 flip questions per pairwise case with stratified dataset quotas and a per-family cap of 600, yielding 1{,}300 questions spanning eight datasets and all \nfamilies families after filtering degenerate responses (empty or single non-letter characters). Each question, along with its associated image(s), is independently classified (temperature = 0) into one of three failure categories. The annotator prompt is blinded to model identities (models are labeled ``Model~A'' and ``Model~B'') to reduce bias. As a secondary check, we also run Claude Opus~4.6 with the same prompt; the two judges agree on 90.8\% of questions (Cohen's $\kappa = 0.30$). Full prompts and API model identifiers are provided in \cref{app:prompts}.

To validate the LLM annotations, three authors independently labeled a stratified subsample of 127 questions using the same three categories. Human inter-annotator agreement was substantial (Fleiss $\kappa = 0.65$, pairwise agreement 87--92\%). The human majority vote classified 84.3\% as genuine errors and 15.7\% as bad annotation or ambiguous, confirming that humans identify more annotation issues than either LLM judge (GPT: 2.6\%, Claude: 10.5\%). The majority vote agreed with GPT on 88.2\% and with Claude on 84.3\% of questions (full breakdown in \cref{app:human_annotation}).

Of the 1{,}300 sampled flips, GPT-5.4 classified 97.4\% (1{,}266) as genuine errors, 2.6\% (34) as bad annotation or ambiguous, and none as format mismatch. All five raters (three human, two LLM) classify the majority as genuine errors, though with different thresholds: GPT-5.4 labels 97.4\% as genuine, while the human majority vote applies a stricter standard, classifying 84.3\% as genuine and flagging 15.7\% as bad annotation or ambiguous. We use the human rate as our conservative estimate: at least 84\% of inverse scaling flips reflect real model failures. The human-flagged bad-annotation cases are concentrated in MME (known annotation issues) and knowledge-heavy benchmarks where ground-truth answers are debatable; on visual reasoning benchmarks (BLINK, ScienceQA), both humans and LLMs agree that nearly all flips are genuine errors.

\section{Discussion}
\label{sec:discussion}

\paragraph{Why does inverse scaling occur in VLMs?}
While our study does not establish causal explanations, several hypotheses fit the observed patterns. First, \emph{localized capability regression}: on some perception-centric subtasks (BLINK, MME), larger released variants are less accurate than smaller ones, even though the larger model is ahead in 74.8\% of eligible comparisons. These regressions affect particular skills, not whole benchmarks, and our data cannot attribute them to parameter count alone. Second, \emph{sensitivity to distractors}: on knowledge-heavy benchmarks (MMMU, MMT-Bench), larger models may be more susceptible to plausible but incorrect options, consistent with the distractor memorization hypothesis of \citet{mckenzie2023inverse}. Third, \emph{MoE-specific factors}: MoE variants frequently appear in inverse scaling pairs, though routing, lower active parameter counts, and training differences cannot be separated in our observational data. Fourth, \emph{training data composition}: larger models typically see more training data, whose composition may introduce biases that hurt specific subtasks.

\paragraph{Can chain-of-thought prompting recover the errors?}
To test whether inverse scaling errors reflect shallow reasoning deficits, we re-evaluated both models of every case under chain-of-thought (CoT) prompting on all shared questions (66{,}698 matched question instances across 375 cases). The CoT prompt adds a step-by-step instruction to the direct prompt (\cref{app:cot}); decoding is unchanged. CoT attenuates inverse scaling but does not remove it: the median Net falls from $+8$ to $0$, yet 96 of 375 cases (25.6\%) retain Net~$\geq 5$, 101 reverse, and the larger model recovers 43.5\% of the questions it had lost under direct prompting. Persistence varies by benchmark, from 48.6\% of MME and 42.1\% of BLINK cases to 7.6\% on MMMU test (full results in \cref{app:cot}). Because the smaller model is also re-evaluated, the persisting cases cannot be attributed to the smaller model reasoning where the larger one does not. CoT thus corrects fewer than half of the individual errors and leaves a quarter of the cases intact, and part of the reduction in Net comes from the smaller model losing accuracy under CoT (in 251 of 375 cases) rather than from the larger model gaining. It also remains an open question whether CoT recovery reflects genuine improvements in reasoning, as recent work suggests that intermediate steps can sometimes mask underlying vulnerabilities~\citep{xu2025visulogic, kambhampati2025stop, shojaee2025illusion}.

\paragraph{Implications for model selection.}
Our results have practical implications. The common practice of selecting the largest available model is suboptimal for some task categories. On a minority of evaluated subtask--model-pair comparisons, concentrated in multi-image spatial reasoning, visual perception, and specific STEM subtasks, smaller models outperform their larger counterparts. Practitioners should evaluate candidate models on their downstream tasks, not on aggregate scores.

\paragraph{Limitations.}
Our study has several limitations. First, we evaluate only open-weight models, since the unknown architectures and parameter counts of proprietary models make scaling analysis unverifiable. Second, the flip analysis weights all questions equally, regardless of difficulty. Third, cross-family replication reports empirical evidence and does not establish any causal mechanism. Fourth, models released under one family name may differ in vision encoder (\cref{app:encoders}), training data, alignment, and active parameters, so our results describe subtask-level regressions among released variants rather than causal effects of parameter count alone; a controlled study training VLMs of several sizes under one fully released recipe would isolate parameter count and make such analyses reproducible, which we leave to future work. Fifth, larger checkpoints may have seen more benchmark data, and we cannot rule out size-dependent contamination. Sixth, results depend on option order; the strongest BLINK cases survive reorderings (\cref{app:option_order}), but paraphrases are untested.
%
\section{Conclusion}
\label{sec:conclusion}
To our knowledge, this is the first systematic study of inverse scaling in vision-language models. We evaluate \nmodels models across \nfamilies families and eight benchmarks. Through per-question pairwise analysis, we identify \ncases cases in which smaller models outperform larger ones, with \ncrossfamily subtask types showing this pattern across multiple families. Our findings demonstrate that VLM performance does not scale monotonically at the subtask level, with multi-image spatial reasoning, visual perception, and specialized STEM knowledge most affected. These results motivate work on the mechanisms of inverse scaling and suggest task-aware model deployment. Finally, this work complements recent evidence of deficiencies in current VLMs (e.g., \citet{asadi2026mirage}), highlighting the need for robust multimodal models that leverage their full capacity across modalities.
\subsection*{AI use statement}

In this work, we used generative AI tools as a writing aid to improve fluency and check grammar, and for coding assistance. We have reviewed all AI-assisted work; AI-assisted code was verified by the authors. We take responsibility for the final content of this work, including text, claims or artifacts produced with the aid of generative AI.

\subsection*{Acknowledgments}
Rohit Saxena was supported by the Engineering and Physical Sciences Research Council (EPSRC) through the AI Hub in Generative Models (grant number EP/Y028805/1).
Pasquale Minervini was partially funded by ELIAI (The Edinburgh Laboratory for Integrated Artificial Intelligence), EPSRC (grant no.\ EP/W002876/1), and a donation from Accenture LLP.
This work was supported by the Edinburgh International Data Facility (EIDF) and the Data-Driven Innovation Programme at the University of Edinburgh.
\subsection*{Reproducibility statement}
%
All 19 evaluated models are open-weight; their exact HuggingFace identifiers are listed in Appendix~\ref{app:details} (Table~\ref{tab:model_ids}), which also states the compute used and the API models that support evaluation. Generation settings, including those of the chain-of-thought control, are given in Appendix~\ref{app:generation}; the answer-verification pipeline and the case-selection thresholds are given in Section~\ref{sec:protocol}. The exact prompts for every benchmark and the annotation prompts for the error-taxonomy judges are reproduced in Appendix~\ref{app:prompts}. Benchmark splits and sizes are listed in Table~\ref{tab:setup}.

\bibliography{main}
\bibliographystyle{iclr2027_conference}

\appendix
\crefalias{section}{appendix}
\section{Additional Experiment Details}
\label{app:details}

\paragraph{Model identifiers.} All models are open-weight and publicly available on HuggingFace. \cref{tab:model_ids} lists the exact HuggingFace model identifiers used for inference.

\begin{table}[h]
\caption{HuggingFace model identifiers for all 19 evaluated models.}
\label{tab:model_ids}
\centering
\small
\begin{tabular}{lll}
\toprule
\textbf{Family} & \textbf{Size} & \textbf{HuggingFace Identifier} \\
\midrule
\multirow{6}{*}{Qwen3-VL} & 2B & \texttt{Qwen/Qwen3-VL-2B-Instruct} \\
& 4B & \texttt{Qwen/Qwen3-VL-4B-Instruct} \\
& 8B & \texttt{Qwen/Qwen3-VL-8B-Instruct} \\
& 30B-A3B & \texttt{Qwen/Qwen3-VL-30B-A3B-Instruct} \\
& 32B & \texttt{Qwen/Qwen3-VL-32B-Instruct} \\
& 235B-A22B & \texttt{Qwen/Qwen3-VL-235B-A22B-Instruct} \\
\midrule
\multirow{8}{*}{InternVL3.5} & 1B & \texttt{OpenGVLab/InternVL3\_5-1B-Instruct} \\
& 2B & \texttt{OpenGVLab/InternVL3\_5-2B-Instruct} \\
& 4B & \texttt{OpenGVLab/InternVL3\_5-4B-Instruct} \\
& 8B & \texttt{OpenGVLab/InternVL3\_5-8B-Instruct} \\
& 14B & \texttt{OpenGVLab/InternVL3\_5-14B-Instruct} \\
& 30B-A3B & \texttt{OpenGVLab/InternVL3\_5-30B-A3B-Instruct} \\
& 38B & \texttt{OpenGVLab/InternVL3\_5-38B-Instruct} \\
& 241B-A28B & \texttt{OpenGVLab/InternVL3\_5-241B-A28B-Instruct} \\
\midrule
\multirow{3}{*}{Gemma~3} & 4B & \texttt{google/gemma-3-4b-it} \\
& 12B & \texttt{google/gemma-3-12b-it} \\
& 27B & \texttt{google/gemma-3-27b-it} \\
\midrule
\multirow{2}{*}{Molmo~2} & 4B & \texttt{allenai/Molmo2-4B} \\
& 8B & \texttt{allenai/Molmo2-8B} \\
\bottomrule
\end{tabular}
\end{table}

\paragraph{Compute.} All inference was performed on NVIDIA H100 GPUs with no quantization. The total evaluation comprises 171 inference runs (19 models $\times$ 9 evaluation splits), i.e., 558{,}334 model--question evaluations.

\paragraph{API models.} Two proprietary models are used for evaluation support, not as evaluation targets:
\begin{itemize}[nosep,leftmargin=*]
    \item \textbf{GPT-5.4} (\texttt{gpt-5.4} via OpenAI API): used as the multimodal fallback adjudicator for answer verification (0.4\% of responses) and as the primary error taxonomy annotator (\cref{sec:taxonomy}).
    \item \textbf{Claude Opus~4.6} (\texttt{claude-opus-4-6} via Anthropic API): used as the secondary error taxonomy annotator for inter-judge agreement analysis.
\end{itemize}

\paragraph{Generation settings.}\label{app:generation} All models are evaluated zero-shot with direct prompting and greedy decoding (\texttt{do\_sample=False}, \texttt{max\_new\_tokens=2048}), with thinking modes disabled. For multiple-choice questions the prompt instructs the model to respond with only the letter of the correct option; for open-ended questions (MMMU, MathVista) it asks for a concise answer. The chain-of-thought control (Appendix~\ref{app:cot}) also uses greedy decoding, with \texttt{max\_new\_tokens=4096}. Prompts are listed in Appendix~\ref{app:prompts}.

\paragraph{Answer verification.} We first apply deterministic answer extraction and matching: answer-letter extraction for multiple-choice questions and exact or numeric matching for open-ended responses. When these rules are inconclusive, GPT-5.4 (OpenAI API, model identifier \texttt{gpt-5.4}) adjudicates as a multimodal fallback, receiving the image, question, options, and model response (prompt in Appendix~\ref{app:prompts}). Across all models and datasets, 0.4\% of responses require adjudication: 4.1\% on BLINK, 2.9\% on MathVista, under 0.05\% on AI2D, MMT-Bench, ScienceQA, and MMMU, and none on MME and MMBench.

\section{Aggregate Accuracy Tables}
\label{app:aggregate}

\begin{table}[h]
\caption{Overall accuracy (\%) for Qwen3-VL. Bold = best in family.}
\centering
\small
\setlength{\tabcolsep}{3pt}
\begin{tabular}{l|cccccc}
\toprule
\textbf{Dataset} & \textbf{2B} & \textbf{4B} & \textbf{8B} & \textbf{30B-A3B} & \textbf{32B} & \textbf{235B-A22B} \\
\midrule
MMT-Bench & 60.2 & 67.3 & 68.5 & 69.7 & 71.7 & \textbf{73.3} \\
MMMU dev\_val & 44.0 & 51.0 & 53.1 & 59.6 & 61.7 & \textbf{67.3} \\
MMMU test & 41.6 & 48.7 & 51.0 & 56.0 & 57.0 & \textbf{64.2} \\
ScienceQA & 87.3 & 92.6 & 94.4 & 95.7 & 96.2 & \textbf{98.0} \\
MME & 82.5 & 87.2 & 89.6 & 90.4 & 90.9 & \textbf{91.6} \\
AI2D & 76.6 & 83.4 & 85.1 & 87.3 & 87.9 & \textbf{89.5} \\
MathVista & 57.2 & 70.9 & 73.6 & 74.5 & 79.4 & \textbf{81.5} \\
BLINK & 54.7 & 66.4 & 65.9 & 62.4 & 66.7 & \textbf{68.8} \\
MMBench & 84.1 & 88.6 & 89.8 & 91.7 & 91.7 & \textbf{92.5} \\
\bottomrule
\end{tabular}
\end{table}

\begin{table}[h]
\caption{Overall accuracy (\%) for InternVL3.5. Bold = best in family.}
\centering
\small
\setlength{\tabcolsep}{2.5pt}
\begin{tabular}{l|cccccccc}
\toprule
\textbf{Dataset} & \textbf{1B} & \textbf{2B} & \textbf{4B} & \textbf{8B} & \textbf{14B} & \textbf{30B-A3B} & \textbf{38B} & \textbf{241B-A28B} \\
\midrule
MMT-Bench & 53.1 & 58.9 & 65.2 & 67.6 & 68.4 & 67.3 & 71.9 & \textbf{72.6} \\
MMMU dev\_val & 40.4 & 47.4 & 57.1 & 57.0 & 59.3 & 61.6 & 63.4 & \textbf{65.2} \\
MMMU test & 37.6 & 44.0 & 50.3 & 52.7 & 54.0 & 55.4 & 57.6 & \textbf{63.2} \\
ScienceQA & 84.1 & 91.7 & 97.3 & 97.5 & 98.3 & 99.2 & 98.8 & \textbf{99.5} \\
MME & 77.7 & 79.2 & 84.0 & 88.3 & 85.9 & 87.6 & 88.7 & \textbf{90.0} \\
AI2D & 70.5 & 78.0 & 82.4 & 83.3 & 84.5 & 86.4 & \textbf{87.1} & \textbf{87.1} \\
MathVista & 48.4 & 61.7 & 67.2 & 69.8 & 70.7 & 72.9 & 74.9 & \textbf{76.7} \\
BLINK & 39.9 & 44.5 & 46.7 & 47.4 & 49.5 & 48.3 & 55.3 & \textbf{57.9} \\
MMBench & 77.9 & 83.8 & 86.9 & 87.8 & 88.7 & 89.5 & 91.2 & \textbf{91.7} \\
\bottomrule
\end{tabular}
\end{table}

\begin{table}[h]
\caption{Overall accuracy (\%) for Gemma~3 and Molmo~2. Bold = best in family. Underline = largest model is not best.}
\centering
\small
\begin{tabular}{l|ccc|cc}
\toprule
\textbf{Dataset} & \multicolumn{3}{c|}{\textit{Gemma~3}} & \multicolumn{2}{c}{\textit{Molmo~2}} \\
& \textbf{4B} & \textbf{12B} & \textbf{27B} & \textbf{4B} & \textbf{8B} \\
\midrule
MMT-Bench & 54.3 & 60.1 & \textbf{61.5} & 64.0 & \textbf{66.5} \\
MMMU dev\_val & 40.7 & 47.9 & \textbf{54.8} & 50.5 & \textbf{52.9} \\
MMMU test & 39.3 & 47.4 & \textbf{50.6} & 46.3 & \textbf{48.0} \\
ScienceQA & 78.5 & 82.2 & \textbf{82.7} & 97.8 & \textbf{98.1} \\
MME & 79.1 & 83.4 & \textbf{85.3} & 76.8 & \textbf{79.3} \\
AI2D & 73.5 & 81.5 & \textbf{83.6} & 85.7 & \textbf{85.9} \\
MathVista & 48.0 & 57.1 & \textbf{63.0} & 57.9 & \textbf{64.2} \\
BLINK & 43.6 & 53.7 & \textbf{57.4} & \textbf{63.5} & \underline{59.3} \\
MMBench & 80.0 & 84.1 & \textbf{86.7} & \textbf{88.3} & \underline{87.9} \\
\bottomrule
\end{tabular}
\end{table}

\begin{table}[h]
\caption{Dense-only cross-family replication: subtask--dataset pairs with inverse scaling in at least three of the four families using only dense model pairs.}
\label{tab:dense_cross}
\centering
\small
\setlength{\tabcolsep}{3pt}
\begin{tabular}{llcc}
\toprule
\textbf{Dataset} & \textbf{Subtask} & \textbf{Families} & \textbf{Max Net} \\
\midrule
BLINK & Multi-view Reasoning & 4 & 30 \\
BLINK & Jigsaw & 3 & 33 \\
MMMU test & Organic Chemistry & 3 & 15 \\
MME & Scene & 3 & 14 \\
MMMU test & Investment & 3 & 13 \\
BLINK & Object Localization & 3 & 11 \\
MMT-Bench & Relation Reasoning & 3 & 9 \\
MMT-Bench & Autonomous Driving & 3 & 8 \\
\bottomrule
\end{tabular}
\end{table}

\section{Full Pairwise Inverse Scaling Results}
\label{app:pairwise}

Tables below list all \ncases pairwise cases identified by the flip analysis (\cref{sec:method}): Net $\geq 5$ on $n \geq 20$ matched questions. Cases are sorted by Net within each dataset. Columns: \textbf{Subtask}: the dataset category; \textbf{Small}/\textbf{Large}: the model pair; $n$: matched questions; \textbf{Acc $\Delta$}: how many percentage points the smaller model's accuracy exceeds the larger model's; \textbf{FI}: FlipInv (smaller correct, larger wrong); \textbf{FN}: FlipNorm (reverse); \textbf{Net}: FI $-$ FN.

\subsection{Qwen3-VL (122 cases)}

\paragraph{MMT-Bench (31).}

\small
\setlength{\tabcolsep}{3pt}
\begin{longtable}{l l l r r r r r}
\toprule
Subtask & Small & Large & $n$ & Acc $\Delta$ & FI & FN & Net \\
\midrule
\endfirsthead
\toprule
Subtask & Small & Large & $n$ & Acc $\Delta$ & FI & FN & Net \\
\midrule
\endhead
\midrule
\multicolumn{8}{r}{\textit{continued}} \\
\endfoot
\bottomrule
\endlastfoot
autonomous\_driving & 30B-A3B & 235B-A22B & 100 & 23.0 & 23 & 0 & 23 \\
autonomous\_driving & 2B & 235B-A22B & 100 & 19.0 & 23 & 4 & 19 \\
autonomous\_driving & 8B & 235B-A22B & 100 & 17.0 & 18 & 1 & 17 \\
autonomous\_driving & 32B & 235B-A22B & 100 & 16.0 & 17 & 1 & 16 \\
autonomous\_driving & 4B & 235B-A22B & 100 & 14.0 & 16 & 2 & 14 \\
cross\_image\_matching & 8B & 30B-A3B & 60 & 23.3 & 21 & 7 & 14 \\
medical\_understanding & 4B & 8B & 100 & 13.0 & 14 & 1 & 13 \\
medical\_understanding & 4B & 32B & 100 & 13.0 & 16 & 3 & 13 \\
medical\_understanding & 4B & 235B-A22B & 100 & 13.0 & 17 & 4 & 13 \\
medical\_understanding & 4B & 30B-A3B & 100 & 12.0 & 16 & 4 & 12 \\
visual\_illusion & 30B-A3B & 235B-A22B & 79 & 15.2 & 17 & 5 & 12 \\
anomaly\_detection & 4B & 30B-A3B & 100 & 11.0 & 21 & 10 & 11 \\
anomaly\_detection & 8B & 30B-A3B & 100 & 10.0 & 18 & 8 & 10 \\
temporal\_understanding & 8B & 30B-A3B & 99 & 10.1 & 23 & 13 & 10 \\
temporal\_understanding & 4B & 30B-A3B & 99 & 9.1 & 21 & 12 & 9 \\
cross\_image\_matching & 4B & 30B-A3B & 60 & 13.3 & 14 & 6 & 8 \\
visual\_illusion & 32B & 235B-A22B & 79 & 10.1 & 11 & 3 & 8 \\
autonomous\_driving & 30B-A3B & 32B & 100 & 7.0 & 8 & 1 & 7 \\
visual\_illusion & 2B & 235B-A22B & 79 & 7.6 & 22 & 16 & 6 \\
action\_recognition & 4B & 8B & 100 & 6.0 & 10 & 4 & 6 \\
temporal\_understanding & 8B & 32B & 99 & 6.1 & 14 & 8 & 6 \\
autonomous\_driving & 2B & 4B & 100 & 5.0 & 13 & 8 & 5 \\
multiple\_image\_analysis & 2B & 8B & 40 & 12.5 & 9 & 4 & 5 \\
visual\_illusion & 2B & 8B & 79 & 6.3 & 23 & 18 & 5 \\
emotion & 4B & 8B & 120 & 4.2 & 10 & 5 & 5 \\
relation\_reasoning & 4B & 8B & 80 & 6.2 & 10 & 5 & 5 \\
action\_recognition & 4B & 30B-A3B & 100 & 5.0 & 10 & 5 & 5 \\
temporal\_understanding & 4B & 32B & 99 & 5.1 & 16 & 11 & 5 \\
action\_recognition & 4B & 235B-A22B & 100 & 5.0 & 14 & 9 & 5 \\
pixel\_level\_perception & 8B & 30B-A3B & 114 & 4.4 & 20 & 15 & 5 \\
gui\_navigation & 30B-A3B & 32B & 80 & 6.2 & 14 & 9 & 5 \\
\end{longtable}
\normalsize

\paragraph{MMMU test (34).}

\small
\setlength{\tabcolsep}{3pt}
\begin{longtable}{l l l r r r r r}
\toprule
Subtask & Small & Large & $n$ & Acc $\Delta$ & FI & FN & Net \\
\midrule
\endfirsthead
\toprule
Subtask & Small & Large & $n$ & Acc $\Delta$ & FI & FN & Net \\
\midrule
\endhead
\midrule
\multicolumn{8}{r}{\textit{continued}} \\
\endfoot
\bottomrule
\endlastfoot
Surveying and Mapping & 8B & 30B-A3B & 412 & 4.1 & 63 & 46 & 17 \\
Electrical Circuit & 30B-A3B & 32B & 128 & 13.3 & 24 & 7 & 17 \\
Cost Management & 30B-A3B & 32B & 58 & 24.1 & 17 & 3 & 14 \\
Data Structure and Algorithm & 4B & 30B-A3B & 92 & 12.0 & 18 & 7 & 11 \\
Cost Management & 30B-A3B & 235B-A22B & 58 & 19.0 & 15 & 4 & 11 \\
Data Structure and Algorithm & 8B & 30B-A3B & 92 & 9.8 & 17 & 8 & 9 \\
Engineering Graphics & 8B & 30B-A3B & 89 & 10.1 & 19 & 10 & 9 \\
Physiology & 2B & 8B & 73 & 11.0 & 16 & 8 & 8 \\
Calculus & 4B & 8B & 161 & 5.0 & 23 & 15 & 8 \\
Financial Accounting & 30B-A3B & 32B & 107 & 7.5 & 20 & 12 & 8 \\
Physiology & 2B & 4B & 73 & 9.6 & 14 & 7 & 7 \\
Control System & 4B & 30B-A3B & 45 & 15.6 & 10 & 3 & 7 \\
Control System & 8B & 30B-A3B & 45 & 15.6 & 10 & 3 & 7 \\
AI Content Detection & 2B & 4B & 20 & 30.0 & 9 & 3 & 6 \\
Cost Management & 2B & 8B & 58 & 10.3 & 8 & 2 & 6 \\
Artificial Intelligence & 4B & 8B & 43 & 14.0 & 11 & 5 & 6 \\
Microbiology & 4B & 8B & 60 & 10.0 & 11 & 5 & 6 \\
Operation Research & 4B & 8B & 49 & 12.2 & 10 & 4 & 6 \\
Principles of Management & 4B & 8B & 41 & 14.6 & 9 & 3 & 6 \\
Financial Management & 30B-A3B & 32B & 36 & 16.7 & 9 & 3 & 6 \\
Operating System & 30B-A3B & 32B & 54 & 11.1 & 12 & 6 & 6 \\
Structural Engineering & 30B-A3B & 32B & 94 & 6.4 & 13 & 7 & 6 \\
Graph Theory & 2B & 4B & 78 & 6.4 & 15 & 10 & 5 \\
Investment & 4B & 8B & 119 & 4.2 & 19 & 14 & 5 \\
Surveying and Mapping & 4B & 30B-A3B & 412 & 1.2 & 54 & 49 & 5 \\
Thermodynamics & 8B & 30B-A3B & 163 & 3.1 & 22 & 17 & 5 \\
Engineering Graphics & 8B & 32B & 89 & 5.6 & 20 & 15 & 5 \\
Surveying and Mapping & 8B & 32B & 412 & 1.2 & 46 & 41 & 5 \\
Civil Engineering & 8B & 235B-A22B & 41 & 12.2 & 11 & 6 & 5 \\
Calculus & 30B-A3B & 32B & 161 & 3.1 & 17 & 12 & 5 \\
Design History & 30B-A3B & 32B & 84 & 6.0 & 7 & 2 & 5 \\
Graph Theory & 30B-A3B & 32B & 78 & 6.4 & 13 & 8 & 5 \\
Human Geography & 30B-A3B & 32B & 47 & 10.6 & 7 & 2 & 5 \\
Electrical Circuit & 30B-A3B & 235B-A22B & 128 & 3.9 & 15 & 10 & 5 \\
\end{longtable}
\normalsize

\paragraph{ScienceQA (7).}

\small
\setlength{\tabcolsep}{3pt}
\begin{longtable}{l l l r r r r r}
\toprule
Subtask & Small & Large & $n$ & Acc $\Delta$ & FI & FN & Net \\
\midrule
\endfirsthead
\toprule
Subtask & Small & Large & $n$ & Acc $\Delta$ & FI & FN & Net \\
\midrule
\endhead
\midrule
\multicolumn{8}{r}{\textit{continued}} \\
\endfoot
\bottomrule
\endlastfoot
Velocity, acceleration, and forces & 30B-A3B & 32B & 62 & 14.5 & 12 & 3 & 9 \\
The Americas: geography & 2B & 4B & 45 & 17.8 & 14 & 6 & 8 \\
Velocity, acceleration, and forces & 8B & 32B & 62 & 12.9 & 11 & 3 & 8 \\
Velocity, acceleration, and forces & 4B & 32B & 62 & 9.7 & 8 & 2 & 6 \\
The Americas: geography & 30B-A3B & 32B & 45 & 13.3 & 7 & 1 & 6 \\
Ecosystems & 4B & 8B & 73 & 6.8 & 6 & 1 & 5 \\
Astronomy & 30B-A3B & 32B & 36 & 13.9 & 6 & 1 & 5 \\
\end{longtable}
\normalsize

\paragraph{MME (10).}

\small
\setlength{\tabcolsep}{3pt}
\begin{longtable}{l l l r r r r r}
\toprule
Subtask & Small & Large & $n$ & Acc $\Delta$ & FI & FN & Net \\
\midrule
\endfirsthead
\toprule
Subtask & Small & Large & $n$ & Acc $\Delta$ & FI & FN & Net \\
\midrule
\endhead
\midrule
\multicolumn{8}{r}{\textit{continued}} \\
\endfoot
\bottomrule
\endlastfoot
scene & 2B & 235B-A22B & 400 & 2.8 & 20 & 9 & 11 \\
scene & 32B & 235B-A22B & 400 & 2.8 & 18 & 7 & 11 \\
scene & 30B-A3B & 235B-A22B & 400 & 2.5 & 14 & 4 & 10 \\
scene & 4B & 235B-A22B & 400 & 1.5 & 14 & 8 & 6 \\
celebrity & 8B & 32B & 340 & 1.8 & 20 & 14 & 6 \\
scene & 8B & 235B-A22B & 400 & 1.5 & 13 & 7 & 6 \\
position & 30B-A3B & 235B-A22B & 60 & 10.0 & 7 & 1 & 6 \\
scene & 2B & 4B & 400 & 1.2 & 14 & 9 & 5 \\
scene & 2B & 8B & 400 & 1.2 & 18 & 13 & 5 \\
posters & 32B & 235B-A22B & 294 & 1.7 & 9 & 4 & 5 \\
\end{longtable}
\normalsize

\paragraph{MathVista (1).}

\small
\setlength{\tabcolsep}{3pt}
\begin{longtable}{l l l r r r r r}
\toprule
Subtask & Small & Large & $n$ & Acc $\Delta$ & FI & FN & Net \\
\midrule
\endfirsthead
\toprule
Subtask & Small & Large & $n$ & Acc $\Delta$ & FI & FN & Net \\
\midrule
\endhead
\midrule
\multicolumn{8}{r}{\textit{continued}} \\
\endfoot
\bottomrule
\endlastfoot
math word problem & 8B & 30B-A3B & 186 & 3.8 & 15 & 8 & 7 \\
\end{longtable}
\normalsize

\paragraph{BLINK (37).}

\small
\setlength{\tabcolsep}{3pt}
\begin{longtable}{l l l r r r r r}
\toprule
Subtask & Small & Large & $n$ & Acc $\Delta$ & FI & FN & Net \\
\midrule
\endfirsthead
\toprule
Subtask & Small & Large & $n$ & Acc $\Delta$ & FI & FN & Net \\
\midrule
\endhead
\midrule
\multicolumn{8}{r}{\textit{continued}} \\
\endfoot
\bottomrule
\endlastfoot
Jigsaw & 4B & 30B-A3B & 150 & 24.0 & 45 & 9 & 36 \\
Jigsaw & 4B & 32B & 150 & 22.0 & 44 & 11 & 33 \\
Multi-view Reasoning & 8B & 32B & 133 & 22.6 & 38 & 8 & 30 \\
Multi-view Reasoning & 2B & 32B & 133 & 21.8 & 37 & 8 & 29 \\
Object Localization & 4B & 30B-A3B & 122 & 18.0 & 35 & 13 & 22 \\
Multi-view Reasoning & 4B & 32B & 133 & 16.5 & 37 & 15 & 22 \\
Forensic Detection & 4B & 32B & 132 & 15.2 & 25 & 5 & 20 \\
Forensic Detection & 4B & 235B-A22B & 132 & 15.2 & 27 & 7 & 20 \\
Jigsaw & 4B & 235B-A22B & 150 & 13.3 & 30 & 10 & 20 \\
Multi-view Reasoning & 30B-A3B & 32B & 133 & 15.0 & 39 & 19 & 20 \\
Jigsaw & 4B & 8B & 150 & 12.7 & 27 & 8 & 19 \\
IQ Test & 8B & 30B-A3B & 150 & 11.3 & 25 & 8 & 17 \\
Jigsaw & 8B & 30B-A3B & 150 & 11.3 & 34 & 17 & 17 \\
Relative Reflectance & 32B & 235B-A22B & 134 & 12.7 & 23 & 6 & 17 \\
Forensic Detection & 4B & 30B-A3B & 132 & 12.1 & 23 & 7 & 16 \\
Forensic Detection & 8B & 32B & 132 & 10.6 & 25 & 11 & 14 \\
Jigsaw & 8B & 32B & 150 & 9.3 & 34 & 20 & 14 \\
Forensic Detection & 8B & 235B-A22B & 132 & 10.6 & 26 & 12 & 14 \\
Multi-view Reasoning & 8B & 235B-A22B & 133 & 10.5 & 32 & 18 & 14 \\
Multi-view Reasoning & 2B & 235B-A22B & 133 & 9.8 & 35 & 22 & 13 \\
Object Localization & 4B & 8B & 122 & 9.0 & 28 & 17 & 11 \\
Object Localization & 8B & 30B-A3B & 122 & 9.0 & 28 & 17 & 11 \\
Object Localization & 4B & 32B & 122 & 8.2 & 30 & 20 & 10 \\
Forensic Detection & 8B & 30B-A3B & 132 & 7.6 & 24 & 14 & 10 \\
Multi-view Reasoning & 8B & 30B-A3B & 133 & 7.5 & 22 & 12 & 10 \\
IQ Test & 8B & 32B & 150 & 6.7 & 20 & 10 & 10 \\
Multi-view Reasoning & 2B & 30B-A3B & 133 & 6.8 & 23 & 14 & 9 \\
Relative Reflectance & 4B & 30B-A3B & 134 & 6.0 & 24 & 16 & 8 \\
Functional Correspondence & 32B & 235B-A22B & 130 & 6.2 & 25 & 17 & 8 \\
Multi-view Reasoning & 2B & 4B & 133 & 5.3 & 19 & 12 & 7 \\
Jigsaw & 2B & 30B-A3B & 150 & 4.7 & 39 & 32 & 7 \\
Relative Depth & 8B & 30B-A3B & 124 & 5.6 & 10 & 3 & 7 \\
Forensic Detection & 4B & 8B & 132 & 4.5 & 16 & 10 & 6 \\
Relative Depth & 4B & 30B-A3B & 124 & 4.8 & 13 & 7 & 6 \\
Multi-view Reasoning & 4B & 235B-A22B & 133 & 4.5 & 30 & 24 & 6 \\
Object Localization & 4B & 235B-A22B & 122 & 4.9 & 22 & 16 & 6 \\
Relative Reflectance & 8B & 30B-A3B & 134 & 4.5 & 29 & 23 & 6 \\
\end{longtable}
\normalsize

\paragraph{MMBench (2).}

{
\small
\setlength{\tabcolsep}{3pt}
\begin{longtable}{l l l r r r r r}
\toprule
Subtask & Small & Large & $n$ & Acc $\Delta$ & FI & FN & Net \\
\midrule
\endfirsthead
\toprule
Subtask & Small & Large & $n$ & Acc $\Delta$ & FI & FN & Net \\
\midrule
\endhead
\midrule
\multicolumn{8}{r}{\textit{continued}} \\
\endfoot
\bottomrule
\endlastfoot
coarse\_perception & 30B-A3B & 32B & 1109 & 1.1 & 38 & 26 & 12 \\
attribute\_reasoning & 32B & 235B-A22B & 699 & 0.9 & 19 & 13 & 6 \\
\end{longtable}
\normalsize
}

\subsection{InternVL3.5 (207 cases)}

\paragraph{MMT-Bench (49).}

\small
\setlength{\tabcolsep}{3pt}
\begin{longtable}{l l l r r r r r}
\toprule
Subtask & Small & Large & $n$ & Acc $\Delta$ & FI & FN & Net \\
\midrule
\endfirsthead
\toprule
Subtask & Small & Large & $n$ & Acc $\Delta$ & FI & FN & Net \\
\midrule
\endhead
\midrule
\multicolumn{8}{r}{\textit{continued}} \\
\endfoot
\bottomrule
\endlastfoot
cross\_image\_matching & 14B & 30B-A3B & 60 & 31.7 & 21 & 2 & 19 \\
localization & 8B & 30B-A3B & 169 & 9.5 & 29 & 13 & 16 \\
localization & 14B & 30B-A3B & 169 & 9.5 & 31 & 15 & 16 \\
image\_retrieval & 14B & 30B-A3B & 140 & 10.0 & 22 & 8 & 14 \\
image\_retrieval & 38B & 241B-A28B & 140 & 10.0 & 19 & 5 & 14 \\
temporal\_understanding & 38B & 241B-A28B & 99 & 13.1 & 20 & 7 & 13 \\
visual\_illusion & 1B & 2B & 79 & 15.2 & 21 & 9 & 12 \\
cross\_image\_matching & 1B & 30B-A3B & 60 & 20.0 & 21 & 9 & 12 \\
counting & 8B & 14B & 140 & 8.6 & 29 & 17 & 12 \\
image-to-image\_translation & 8B & 14B & 40 & 30.0 & 13 & 1 & 12 \\
cross\_image\_matching & 38B & 241B-A28B & 60 & 20.0 & 16 & 4 & 12 \\
counting & 2B & 14B & 140 & 7.9 & 29 & 18 & 11 \\
cross\_image\_matching & 14B & 241B-A28B & 60 & 18.3 & 16 & 5 & 11 \\
multiple\_image\_analysis & 1B & 2B & 40 & 22.5 & 12 & 3 & 9 \\
relation\_reasoning & 1B & 2B & 80 & 11.2 & 17 & 8 & 9 \\
visual\_illusion & 1B & 8B & 79 & 11.4 & 23 & 14 & 9 \\
visual\_illusion & 4B & 8B & 79 & 11.4 & 14 & 5 & 9 \\
image-to-image\_translation & 8B & 30B-A3B & 40 & 22.5 & 12 & 3 & 9 \\
threed & 38B & 241B-A28B & 40 & 22.5 & 11 & 2 & 9 \\
cross\_image\_matching & 2B & 30B-A3B & 60 & 13.3 & 18 & 10 & 8 \\
cross\_image\_matching & 4B & 30B-A3B & 60 & 13.3 & 16 & 8 & 8 \\
image\_retrieval & 8B & 30B-A3B & 140 & 5.7 & 24 & 16 & 8 \\
visual\_recognition & 14B & 30B-A3B & 658 & 1.2 & 48 & 40 & 8 \\
doc\_understanding & 8B & 30B-A3B & 124 & 5.6 & 13 & 6 & 7 \\
ocr & 8B & 30B-A3B & 60 & 11.7 & 9 & 2 & 7 \\
image-to-image\_translation & 8B & 38B & 40 & 17.5 & 11 & 4 & 7 \\
image-to-image\_translation & 8B & 241B-A28B & 40 & 17.5 & 10 & 3 & 7 \\
doc\_understanding & 14B & 30B-A3B & 124 & 5.6 & 13 & 6 & 7 \\
cross\_image\_matching & 1B & 8B & 60 & 10.0 & 15 & 9 & 6 \\
visual\_illusion & 1B & 30B-A3B & 79 & 7.6 & 19 & 13 & 6 \\
visual\_illusion & 1B & 38B & 79 & 7.6 & 22 & 16 & 6 \\
counting & 2B & 4B & 140 & 4.3 & 25 & 19 & 6 \\
visual\_illusion & 4B & 30B-A3B & 79 & 7.6 & 14 & 8 & 6 \\
visual\_illusion & 4B & 38B & 79 & 7.6 & 16 & 10 & 6 \\
cross\_image\_matching & 8B & 30B-A3B & 60 & 10.0 & 12 & 6 & 6 \\
multiple\_image\_analysis & 1B & 8B & 40 & 12.5 & 11 & 6 & 5 \\
autonomous\_driving & 2B & 4B & 100 & 5.0 & 9 & 4 & 5 \\
hallucination & 4B & 8B & 80 & 6.2 & 8 & 3 & 5 \\
counting & 4B & 14B & 140 & 3.6 & 22 & 17 & 5 \\
ocr & 4B & 30B-A3B & 60 & 8.3 & 8 & 3 & 5 \\
ocr & 8B & 14B & 60 & 8.3 & 7 & 2 & 5 \\
counting & 8B & 30B-A3B & 140 & 3.6 & 25 & 20 & 5 \\
counting & 8B & 38B & 140 & 3.6 & 29 & 24 & 5 \\
visual\_captioning & 8B & 38B & 159 & 3.1 & 11 & 6 & 5 \\
image\_evaluation\_judgement & 14B & 30B-A3B & 40 & 12.5 & 7 & 2 & 5 \\
image\_retrieval & 14B & 241B-A28B & 140 & 3.6 & 14 & 9 & 5 \\
temporal\_understanding & 14B & 241B-A28B & 99 & 5.1 & 18 & 13 & 5 \\
emotion & 30B-A3B & 241B-A28B & 120 & 4.2 & 11 & 6 & 5 \\
localization & 38B & 241B-A28B & 169 & 3.0 & 25 & 20 & 5 \\
\end{longtable}
\normalsize

\paragraph{MMMU dev\_val (10).}

\small
\setlength{\tabcolsep}{3pt}
\begin{longtable}{l l l r r r r r}
\toprule
Subtask & Small & Large & $n$ & Acc $\Delta$ & FI & FN & Net \\
\midrule
\endfirsthead
\toprule
Subtask & Small & Large & $n$ & Acc $\Delta$ & FI & FN & Net \\
\midrule
\endhead
\midrule
\multicolumn{8}{r}{\textit{continued}} \\
\endfoot
\bottomrule
\endlastfoot
Mechanics of Materials & 38B & 241B-A28B & 29 & 34.5 & 10 & 0 & 10 \\
Mechanics of Materials & 30B-A3B & 241B-A28B & 29 & 31.0 & 9 & 0 & 9 \\
Music & 8B & 14B & 35 & 22.9 & 11 & 3 & 8 \\
Market Research & 4B & 14B & 35 & 20.0 & 9 & 2 & 7 \\
Mechanics of Materials & 14B & 241B-A28B & 29 & 24.1 & 8 & 1 & 7 \\
Mechanics of Materials & 2B & 241B-A28B & 29 & 20.7 & 10 & 4 & 6 \\
Music & 4B & 14B & 35 & 17.1 & 9 & 3 & 6 \\
Mechanics of Materials & 2B & 8B & 29 & 17.2 & 9 & 4 & 5 \\
Mechanics of Materials & 4B & 241B-A28B & 29 & 17.2 & 8 & 3 & 5 \\
Market Research & 8B & 14B & 35 & 14.3 & 7 & 2 & 5 \\
\end{longtable}
\normalsize

\paragraph{MMMU test (57).}

\small
\setlength{\tabcolsep}{3pt}
\begin{longtable}{l l l r r r r r}
\toprule
Subtask & Small & Large & $n$ & Acc $\Delta$ & FI & FN & Net \\
\midrule
\endfirsthead
\toprule
Subtask & Small & Large & $n$ & Acc $\Delta$ & FI & FN & Net \\
\midrule
\endhead
\midrule
\multicolumn{8}{r}{\textit{continued}} \\
\endfoot
\bottomrule
\endlastfoot
Organic Chemistry & 8B & 14B & 322 & 4.7 & 52 & 37 & 15 \\
Engineering Graphics & 4B & 14B & 89 & 15.7 & 22 & 8 & 14 \\
Music & 38B & 241B-A28B & 334 & 3.9 & 63 & 50 & 13 \\
Organic Chemistry & 4B & 14B & 322 & 3.7 & 51 & 39 & 12 \\
Engineering Graphics & 4B & 38B & 89 & 13.5 & 28 & 16 & 12 \\
Genetics & 14B & 30B-A3B & 57 & 21.1 & 15 & 3 & 12 \\
Investment & 14B & 30B-A3B & 119 & 10.1 & 24 & 12 & 12 \\
Music & 8B & 14B & 334 & 3.3 & 55 & 44 & 11 \\
Probability and Statistics & 8B & 14B & 53 & 20.8 & 13 & 2 & 11 \\
Neuropathology & 4B & 14B & 94 & 10.6 & 14 & 4 & 10 \\
Neuropathology & 4B & 30B-A3B & 94 & 10.6 & 20 & 10 & 10 \\
Electrical Circuit & 8B & 14B & 128 & 7.8 & 18 & 8 & 10 \\
Geometry & 30B-A3B & 38B & 148 & 6.8 & 26 & 16 & 10 \\
Engineering Graphics & 2B & 14B & 89 & 10.1 & 19 & 10 & 9 \\
Artificial Intelligence & 4B & 8B & 43 & 20.9 & 10 & 1 & 9 \\
Investment & 4B & 8B & 119 & 7.6 & 25 & 16 & 9 \\
Engineering Dynamics & 8B & 14B & 65 & 13.8 & 15 & 6 & 9 \\
Structural Engineering & 8B & 14B & 94 & 9.6 & 18 & 9 & 9 \\
Mechanical Design & 1B & 8B & 92 & 8.7 & 21 & 13 & 8 \\
Financial Accounting & 4B & 8B & 107 & 7.5 & 19 & 11 & 8 \\
Genetics & 4B & 30B-A3B & 57 & 14.0 & 15 & 7 & 8 \\
Thermodynamics & 8B & 14B & 163 & 4.9 & 36 & 28 & 8 \\
Managerial Accounting & 14B & 30B-A3B & 154 & 5.2 & 26 & 18 & 8 \\
Engineering Graphics & 30B-A3B & 38B & 89 & 9.0 & 22 & 14 & 8 \\
Fluid Dynamics & 1B & 2B & 46 & 15.2 & 15 & 8 & 7 \\
Mechanical Design & 1B & 4B & 92 & 7.6 & 19 & 12 & 7 \\
Engineering Graphics & 2B & 38B & 89 & 7.9 & 22 & 15 & 7 \\
Engineering Graphics & 4B & 8B & 89 & 7.9 & 16 & 9 & 7 \\
Artificial Intelligence & 4B & 14B & 43 & 16.3 & 8 & 1 & 7 \\
Structural Engineering & 4B & 14B & 94 & 7.4 & 17 & 10 & 7 \\
Artificial Intelligence & 4B & 30B-A3B & 43 & 16.3 & 9 & 2 & 7 \\
Engineering Graphics & 8B & 14B & 89 & 7.9 & 18 & 11 & 7 \\
Graph Theory & 8B & 14B & 78 & 9.0 & 16 & 9 & 7 \\
Probability and Statistics & 8B & 30B-A3B & 53 & 13.2 & 13 & 6 & 7 \\
Plant Pathology & 14B & 30B-A3B & 266 & 2.6 & 32 & 25 & 7 \\
Engineering Dynamics & 2B & 14B & 65 & 9.2 & 14 & 8 & 6 \\
Neuropathology & 4B & 8B & 94 & 6.4 & 16 & 10 & 6 \\
Investment & 4B & 30B-A3B & 119 & 5.0 & 18 & 12 & 6 \\
Calculus & 8B & 14B & 161 & 3.7 & 18 & 12 & 6 \\
Materials Science & 8B & 14B & 139 & 4.3 & 17 & 11 & 6 \\
Compiler Principle & 8B & 30B-A3B & 45 & 13.3 & 10 & 4 & 6 \\
Computer Network & 8B & 30B-A3B & 50 & 12.0 & 10 & 4 & 6 \\
Engineering Dynamics & 8B & 30B-A3B & 65 & 9.2 & 17 & 11 & 6 \\
Organic Chemistry & 8B & 30B-A3B & 322 & 1.9 & 41 & 35 & 6 \\
Structural Engineering & 8B & 30B-A3B & 94 & 6.4 & 17 & 11 & 6 \\
Genetics & 14B & 38B & 57 & 10.5 & 11 & 5 & 6 \\
Analog Electronics & 1B & 2B & 39 & 12.8 & 5 & 0 & 5 \\
Industrial Design & 2B & 8B & 44 & 11.4 & 6 & 1 & 5 \\
Mechanical Design & 2B & 8B & 92 & 5.4 & 17 & 12 & 5 \\
Fluid Dynamics & 8B & 14B & 46 & 10.9 & 10 & 5 & 5 \\
Genetics & 8B & 30B-A3B & 57 & 8.8 & 10 & 5 & 5 \\
Music & 8B & 30B-A3B & 334 & 1.5 & 63 & 58 & 5 \\
Project Management & 8B & 30B-A3B & 25 & 20.0 & 6 & 1 & 5 \\
Engineering Graphics & 8B & 38B & 89 & 5.6 & 22 & 17 & 5 \\
Probability and Statistics & 8B & 38B & 53 & 9.4 & 9 & 4 & 5 \\
Epidemiology & 14B & 30B-A3B & 423 & 1.2 & 61 & 56 & 5 \\
Principals of Economics & 30B-A3B & 38B & 57 & 8.8 & 13 & 8 & 5 \\
\end{longtable}
\normalsize

\paragraph{MME (23).}

\small
\setlength{\tabcolsep}{3pt}
\begin{longtable}{l l l r r r r r}
\toprule
Subtask & Small & Large & $n$ & Acc $\Delta$ & FI & FN & Net \\
\midrule
\endfirsthead
\toprule
Subtask & Small & Large & $n$ & Acc $\Delta$ & FI & FN & Net \\
\midrule
\endhead
\midrule
\multicolumn{8}{r}{\textit{continued}} \\
\endfoot
\bottomrule
\endlastfoot
landmark & 1B & 2B & 400 & 7.2 & 55 & 26 & 29 \\
celebrity & 8B & 14B & 340 & 7.4 & 31 & 6 & 25 \\
landmark & 8B & 14B & 400 & 5.0 & 34 & 14 & 20 \\
scene & 8B & 241B-A28B & 400 & 4.0 & 20 & 4 & 16 \\
landmark & 8B & 30B-A3B & 400 & 3.8 & 35 & 20 & 15 \\
celebrity & 8B & 30B-A3B & 340 & 4.1 & 31 & 17 & 14 \\
scene & 8B & 38B & 400 & 3.5 & 21 & 7 & 14 \\
artwork & 8B & 14B & 400 & 3.2 & 24 & 11 & 13 \\
scene & 2B & 241B-A28B & 400 & 3.0 & 21 & 9 & 12 \\
scene & 2B & 38B & 400 & 2.5 & 21 & 11 & 10 \\
scene & 8B & 14B & 400 & 2.2 & 17 & 8 & 9 \\
scene & 30B-A3B & 241B-A28B & 400 & 2.2 & 17 & 8 & 9 \\
scene & 8B & 30B-A3B & 400 & 1.8 & 17 & 10 & 7 \\
scene & 14B & 241B-A28B & 400 & 1.8 & 18 & 11 & 7 \\
scene & 30B-A3B & 38B & 400 & 1.8 & 17 & 10 & 7 \\
color & 1B & 2B & 60 & 10.0 & 7 & 1 & 6 \\
scene & 2B & 4B & 400 & 1.5 & 15 & 9 & 6 \\
scene & 4B & 241B-A28B & 400 & 1.5 & 16 & 10 & 6 \\
celebrity & 8B & 38B & 340 & 1.8 & 17 & 11 & 6 \\
artwork & 1B & 2B & 400 & 1.2 & 54 & 49 & 5 \\
scene & 2B & 14B & 400 & 1.2 & 17 & 12 & 5 \\
artwork & 4B & 14B & 400 & 1.2 & 25 & 20 & 5 \\
scene & 14B & 38B & 400 & 1.2 & 16 & 11 & 5 \\
\end{longtable}
\normalsize

\paragraph{AI2D (8).}

\small
\setlength{\tabcolsep}{3pt}
\begin{longtable}{l l l r r r r r}
\toprule
Subtask & Small & Large & $n$ & Acc $\Delta$ & FI & FN & Net \\
\midrule
\endfirsthead
\toprule
Subtask & Small & Large & $n$ & Acc $\Delta$ & FI & FN & Net \\
\midrule
\endhead
\midrule
\multicolumn{8}{r}{\textit{continued}} \\
\endfoot
\bottomrule
\endlastfoot
moonPhaseEquinox & 30B-A3B & 38B & 277 & 5.8 & 29 & 13 & 16 \\
moonPhaseEquinox & 30B-A3B & 241B-A28B & 277 & 3.2 & 20 & 11 & 9 \\
photosynthesisRespiration & 14B & 38B & 79 & 7.6 & 8 & 2 & 6 \\
rockCycle & 38B & 241B-A28B & 67 & 9.0 & 9 & 3 & 6 \\
partsOfA & 14B & 30B-A3B & 487 & 1.0 & 33 & 28 & 5 \\
photosynthesisRespiration & 14B & 30B-A3B & 79 & 6.3 & 5 & 0 & 5 \\
foodChainsWebs & 38B & 241B-A28B & 1162 & 0.4 & 38 & 33 & 5 \\
partsOfA & 38B & 241B-A28B & 487 & 1.0 & 31 & 26 & 5 \\
\end{longtable}
\normalsize

\paragraph{MathVista (2).}

\small
\setlength{\tabcolsep}{3pt}
\begin{longtable}{l l l r r r r r}
\toprule
Subtask & Small & Large & $n$ & Acc $\Delta$ & FI & FN & Net \\
\midrule
\endfirsthead
\toprule
Subtask & Small & Large & $n$ & Acc $\Delta$ & FI & FN & Net \\
\midrule
\endhead
\midrule
\multicolumn{8}{r}{\textit{continued}} \\
\endfoot
\bottomrule
\endlastfoot
geometry problem solving & 8B & 14B & 208 & 4.3 & 33 & 24 & 9 \\
visual question answering & 38B & 241B-A28B & 179 & 3.4 & 20 & 14 & 6 \\
\end{longtable}
\normalsize

\paragraph{BLINK (48).}

\small
\setlength{\tabcolsep}{3pt}
\begin{longtable}{l l l r r r r r}
\toprule
Subtask & Small & Large & $n$ & Acc $\Delta$ & FI & FN & Net \\
\midrule
\endfirsthead
\toprule
Subtask & Small & Large & $n$ & Acc $\Delta$ & FI & FN & Net \\
\midrule
\endhead
\midrule
\multicolumn{8}{r}{\textit{continued}} \\
\endfoot
\bottomrule
\endlastfoot
Forensic Detection & 2B & 30B-A3B & 132 & 22.7 & 30 & 0 & 30 \\
Forensic Detection & 4B & 30B-A3B & 132 & 22.0 & 29 & 0 & 29 \\
Forensic Detection & 14B & 30B-A3B & 132 & 22.0 & 29 & 0 & 29 \\
Forensic Detection & 8B & 30B-A3B & 132 & 21.2 & 28 & 0 & 28 \\
Visual Correspondence & 14B & 30B-A3B & 172 & 9.9 & 43 & 26 & 17 \\
Multi-view Reasoning & 1B & 4B & 133 & 12.0 & 52 & 36 & 16 \\
Forensic Detection & 1B & 30B-A3B & 132 & 12.1 & 18 & 2 & 16 \\
Jigsaw & 2B & 30B-A3B & 150 & 10.0 & 38 & 23 & 15 \\
Relative Reflectance & 14B & 38B & 134 & 11.2 & 35 & 20 & 15 \\
Art Style & 38B & 241B-A28B & 117 & 12.8 & 24 & 9 & 15 \\
Relative Reflectance & 14B & 30B-A3B & 134 & 10.4 & 35 & 21 & 14 \\
Art Style & 14B & 30B-A3B & 117 & 11.1 & 36 & 23 & 13 \\
Jigsaw & 2B & 8B & 150 & 8.0 & 27 & 15 & 12 \\
Relative Reflectance & 4B & 38B & 134 & 9.0 & 38 & 26 & 12 \\
Multi-view Reasoning & 30B-A3B & 38B & 133 & 9.0 & 35 & 23 & 12 \\
Multi-view Reasoning & 1B & 14B & 133 & 8.3 & 29 & 18 & 11 \\
Relative Reflectance & 4B & 30B-A3B & 134 & 8.2 & 38 & 27 & 11 \\
Jigsaw & 2B & 14B & 150 & 6.7 & 41 & 31 & 10 \\
Multi-view Reasoning & 8B & 14B & 133 & 7.5 & 35 & 25 & 10 \\
Jigsaw & 4B & 30B-A3B & 150 & 6.0 & 34 & 25 & 9 \\
Art Style & 8B & 30B-A3B & 117 & 7.7 & 32 & 23 & 9 \\
Multi-view Reasoning & 1B & 2B & 133 & 6.0 & 62 & 54 & 8 \\
Multi-view Reasoning & 1B & 38B & 133 & 6.0 & 24 & 16 & 8 \\
Multi-view Reasoning & 2B & 4B & 133 & 6.0 & 25 & 17 & 8 \\
Spatial Relation & 2B & 4B & 143 & 5.6 & 16 & 8 & 8 \\
Art Style & 2B & 30B-A3B & 117 & 6.8 & 35 & 27 & 8 \\
Semantic Correspondence & 8B & 14B & 139 & 5.8 & 26 & 18 & 8 \\
Relative Reflectance & 8B & 38B & 134 & 6.0 & 33 & 25 & 8 \\
Object Localization & 8B & 241B-A28B & 122 & 6.6 & 23 & 15 & 8 \\
Spatial Relation & 2B & 8B & 143 & 4.9 & 14 & 7 & 7 \\
Relative Depth & 4B & 8B & 124 & 5.6 & 15 & 8 & 7 \\
Relative Reflectance & 8B & 30B-A3B & 134 & 5.2 & 35 & 28 & 7 \\
Multi-view Reasoning & 8B & 38B & 133 & 5.3 & 38 & 31 & 7 \\
Multi-view Reasoning & 30B-A3B & 241B-A28B & 133 & 5.3 & 35 & 28 & 7 \\
Object Localization & 30B-A3B & 241B-A28B & 122 & 5.7 & 23 & 16 & 7 \\
Relative Depth & 30B-A3B & 241B-A28B & 124 & 5.6 & 22 & 15 & 7 \\
Art Style & 2B & 4B & 117 & 5.1 & 30 & 24 & 6 \\
Jigsaw & 2B & 4B & 150 & 4.0 & 21 & 15 & 6 \\
Jigsaw & 2B & 38B & 150 & 4.0 & 31 & 25 & 6 \\
Jigsaw & 4B & 8B & 150 & 4.0 & 22 & 16 & 6 \\
Object Localization & 8B & 14B & 122 & 4.9 & 19 & 13 & 6 \\
Relative Reflectance & 14B & 241B-A28B & 134 & 4.5 & 28 & 22 & 6 \\
Relative Depth & 38B & 241B-A28B & 124 & 4.8 & 17 & 11 & 6 \\
Visual Similarity & 38B & 241B-A28B & 135 & 4.4 & 23 & 17 & 6 \\
IQ Test & 1B & 2B & 150 & 3.3 & 31 & 26 & 5 \\
Visual Similarity & 4B & 14B & 135 & 3.7 & 24 & 19 & 5 \\
Jigsaw & 14B & 30B-A3B & 150 & 3.3 & 44 & 39 & 5 \\
Object Localization & 38B & 241B-A28B & 122 & 4.1 & 18 & 13 & 5 \\
\end{longtable}
\normalsize

\paragraph{MMBench (10)}.

{
\small
\setlength{\tabcolsep}{3pt}
\begin{longtable}{l l l r r r r r}
\toprule
Subtask & Small & Large & $n$ & Acc $\Delta$ & FI & FN & Net \\
\midrule
\endfirsthead
\toprule
Subtask & Small & Large & $n$ & Acc $\Delta$ & FI & FN & Net \\
\midrule
\endhead
\midrule
\multicolumn{8}{r}{\textit{continued}} \\
\endfoot
\bottomrule
\endlastfoot
coarse\_perception & 2B & 8B & 1109 & 1.9 & 49 & 28 & 21 \\
attribute\_reasoning & 8B & 14B & 699 & 2.6 & 35 & 17 & 18 \\
coarse\_perception & 2B & 4B & 1109 & 1.5 & 42 & 25 & 17 \\
attribute\_reasoning & 4B & 14B & 699 & 2.3 & 29 & 13 & 16 \\
attribute\_reasoning & 2B & 14B & 699 & 2.1 & 29 & 14 & 15 \\
coarse\_perception & 30B-A3B & 241B-A28B & 1109 & 1.1 & 27 & 15 & 12 \\
finegrained\_perception (cross-instance) & 14B & 30B-A3B & 533 & 1.7 & 36 & 27 & 9 \\
coarse\_perception & 2B & 241B-A28B & 1109 & 0.7 & 39 & 31 & 8 \\
coarse\_perception & 38B & 241B-A28B & 1109 & 0.7 & 32 & 24 & 8 \\
coarse\_perception & 14B & 241B-A28B & 1109 & 0.5 & 34 & 28 & 6 \\
\end{longtable}
\normalsize
}

\subsection{Gemma~3 (26 cases)}

\paragraph{MMT-Bench (6).}

\small
\setlength{\tabcolsep}{3pt}
\begin{longtable}{l l l r r r r r}
\toprule
Subtask & Small & Large & $n$ & Acc $\Delta$ & FI & FN & Net \\
\midrule
\endfirsthead
\toprule
Subtask & Small & Large & $n$ & Acc $\Delta$ & FI & FN & Net \\
\midrule
\endhead
\midrule
\multicolumn{8}{r}{\textit{continued}} \\
\endfoot
\bottomrule
\endlastfoot
visual\_recognition & 12b & 27b & 658 & 2.3 & 37 & 22 & 15 \\
autonomous\_driving & 12b & 27b & 100 & 8.0 & 14 & 6 & 8 \\
medical\_understanding & 4b & 27b & 100 & 7.0 & 20 & 13 & 7 \\
action\_recognition & 4b & 12b & 100 & 6.0 & 12 & 6 & 6 \\
relation\_reasoning & 4b & 12b & 80 & 6.2 & 13 & 8 & 5 \\
threed & 12b & 27b & 40 & 12.5 & 8 & 3 & 5 \\
\end{longtable}
\normalsize

\paragraph{MMMU test (8).}

\small
\setlength{\tabcolsep}{3pt}
\begin{longtable}{l l l r r r r r}
\toprule
Subtask & Small & Large & $n$ & Acc $\Delta$ & FI & FN & Net \\
\midrule
\endfirsthead
\toprule
Subtask & Small & Large & $n$ & Acc $\Delta$ & FI & FN & Net \\
\midrule
\endhead
\midrule
\multicolumn{8}{r}{\textit{continued}} \\
\endfoot
\bottomrule
\endlastfoot
Investment & 4b & 12b & 119 & 10.9 & 31 & 18 & 13 \\
Fluid Mechanics & 12b & 27b & 225 & 4.4 & 49 & 39 & 10 \\
Corporate Finance & 12b & 27b & 131 & 6.9 & 25 & 16 & 9 \\
Pathology & 12b & 27b & 33 & 21.2 & 7 & 0 & 7 \\
Physiology & 4b & 12b & 73 & 8.2 & 16 & 10 & 6 \\
Management Models & 12b & 27b & 72 & 8.3 & 12 & 6 & 6 \\
Mechanics of Materials & 4b & 12b & 364 & 1.4 & 73 & 68 & 5 \\
Organic Chemistry & 12b & 27b & 322 & 1.6 & 43 & 38 & 5 \\
\end{longtable}
\normalsize

\paragraph{ScienceQA (10).}

\small
\setlength{\tabcolsep}{3pt}
\begin{longtable}{l l l r r r r r}
\toprule
Subtask & Small & Large & $n$ & Acc $\Delta$ & FI & FN & Net \\
\midrule
\endfirsthead
\toprule
Subtask & Small & Large & $n$ & Acc $\Delta$ & FI & FN & Net \\
\midrule
\endhead
\midrule
\multicolumn{8}{r}{\textit{continued}} \\
\endfoot
\bottomrule
\endlastfoot
Basic economic principles & 4b & 27b & 71 & 39.4 & 31 & 3 & 28 \\
Particle motion and energy & 12b & 27b & 41 & 43.9 & 18 & 0 & 18 \\
Basic economic principles & 12b & 27b & 71 & 23.9 & 21 & 4 & 17 \\
Particle motion and energy & 4b & 27b & 41 & 34.1 & 19 & 5 & 14 \\
Magnets & 12b & 27b & 110 & 11.8 & 31 & 18 & 13 \\
Basic economic principles & 4b & 12b & 71 & 15.5 & 14 & 3 & 11 \\
Magnets & 4b & 27b & 110 & 10.0 & 36 & 25 & 11 \\
Physical Geography & 4b & 12b & 51 & 11.8 & 6 & 0 & 6 \\
Solutions & 12b & 27b & 54 & 11.1 & 7 & 1 & 6 \\
Oceania: geography & 4b & 12b & 51 & 9.8 & 14 & 9 & 5 \\
\end{longtable}
\normalsize

\paragraph{BLINK (2).}

\small
\setlength{\tabcolsep}{3pt}
\begin{longtable}{l l l r r r r r}
\toprule
Subtask & Small & Large & $n$ & Acc $\Delta$ & FI & FN & Net \\
\midrule
\endfirsthead
\toprule
Subtask & Small & Large & $n$ & Acc $\Delta$ & FI & FN & Net \\
\midrule
\endhead
\midrule
\multicolumn{8}{r}{\textit{continued}} \\
\endfoot
\bottomrule
\endlastfoot
Multi-view Reasoning & 4b & 27b & 133 & 5.3 & 9 & 2 & 7 \\
Multi-view Reasoning & 12b & 27b & 133 & 5.3 & 14 & 7 & 7 \\
\end{longtable}
\normalsize

\subsection{Molmo~2 (20 cases)}

\paragraph{MMT-Bench (1).}

\small
\setlength{\tabcolsep}{3pt}
\begin{longtable}{l l l r r r r r}
\toprule
Subtask & Small & Large & $n$ & Acc $\Delta$ & FI & FN & Net \\
\midrule
\endfirsthead
\toprule
Subtask & Small & Large & $n$ & Acc $\Delta$ & FI & FN & Net \\
\midrule
\endhead
\midrule
\multicolumn{8}{r}{\textit{continued}} \\
\endfoot
\bottomrule
\endlastfoot
emotion & 4B & 8B & 120 & 5.0 & 11 & 5 & 6 \\
\end{longtable}
\normalsize

\paragraph{MMMU test (6).}

\small
\setlength{\tabcolsep}{3pt}
\begin{longtable}{l l l r r r r r}
\toprule
Subtask & Small & Large & $n$ & Acc $\Delta$ & FI & FN & Net \\
\midrule
\endfirsthead
\toprule
Subtask & Small & Large & $n$ & Acc $\Delta$ & FI & FN & Net \\
\midrule
\endhead
\midrule
\multicolumn{8}{r}{\textit{continued}} \\
\endfoot
\bottomrule
\endlastfoot
Materials Science & 4B & 8B & 139 & 11.5 & 24 & 8 & 16 \\
Music & 4B & 8B & 334 & 4.2 & 54 & 40 & 14 \\
Organic Chemistry & 4B & 8B & 322 & 3.7 & 42 & 30 & 12 \\
Thermodynamics & 4B & 8B & 163 & 5.5 & 15 & 6 & 9 \\
Mechanics of Materials & 4B & 8B & 364 & 1.9 & 32 & 25 & 7 \\
Microbiology & 4B & 8B & 60 & 8.3 & 8 & 3 & 5 \\
\end{longtable}
\normalsize

\paragraph{MME (2).}

\small
\setlength{\tabcolsep}{3pt}
\begin{longtable}{l l l r r r r r}
\toprule
Subtask & Small & Large & $n$ & Acc $\Delta$ & FI & FN & Net \\
\midrule
\endfirsthead
\toprule
Subtask & Small & Large & $n$ & Acc $\Delta$ & FI & FN & Net \\
\midrule
\endhead
\midrule
\multicolumn{8}{r}{\textit{continued}} \\
\endfoot
\bottomrule
\endlastfoot
artwork & 4B & 8B & 400 & 4.3 & 45 & 28 & 17 \\
scene & 4B & 8B & 400 & 3.2 & 21 & 8 & 13 \\
\end{longtable}
\normalsize

\paragraph{AI2D (1).}

\small
\setlength{\tabcolsep}{3pt}
\begin{longtable}{l l l r r r r r}
\toprule
Subtask & Small & Large & $n$ & Acc $\Delta$ & FI & FN & Net \\
\midrule
\endfirsthead
\toprule
Subtask & Small & Large & $n$ & Acc $\Delta$ & FI & FN & Net \\
\midrule
\endhead
\midrule
\multicolumn{8}{r}{\textit{continued}} \\
\endfoot
\bottomrule
\endlastfoot
foodChainsWebs & 4B & 8B & 1162 & 0.6 & 36 & 29 & 7 \\
\end{longtable}
\normalsize

\paragraph{BLINK (8).}

\small
\setlength{\tabcolsep}{3pt}
\begin{longtable}{l l l r r r r r}
\toprule
Subtask & Small & Large & $n$ & Acc $\Delta$ & FI & FN & Net \\
\midrule
\endfirsthead
\toprule
Subtask & Small & Large & $n$ & Acc $\Delta$ & FI & FN & Net \\
\midrule
\endhead
\midrule
\multicolumn{8}{r}{\textit{continued}} \\
\endfoot
\bottomrule
\endlastfoot
Jigsaw & 4B & 8B & 150 & 20.7 & 43 & 12 & 31 \\
Visual Correspondence & 4B & 8B & 172 & 9.3 & 29 & 13 & 16 \\
Multi-view Reasoning & 4B & 8B & 133 & 11.3 & 73 & 58 & 15 \\
Art Style & 4B & 8B & 117 & 10.3 & 21 & 9 & 12 \\
Relative Reflectance & 4B & 8B & 134 & 6.7 & 21 & 12 & 9 \\
Semantic Correspondence & 4B & 8B & 139 & 6.5 & 24 & 15 & 9 \\
Forensic Detection & 4B & 8B & 132 & 6.1 & 27 & 19 & 8 \\
Object Localization & 4B & 8B & 122 & 4.1 & 22 & 17 & 5 \\
\end{longtable}
\normalsize

\paragraph{MMBench (2)}.

{
\small
\setlength{\tabcolsep}{3pt}
\begin{longtable}{l l l r r r r r}
\toprule
Subtask & Small & Large & $n$ & Acc $\Delta$ & FI & FN & Net \\
\midrule
\endfirsthead
\toprule
Subtask & Small & Large & $n$ & Acc $\Delta$ & FI & FN & Net \\
\midrule
\endhead
\midrule
\multicolumn{8}{r}{\textit{continued}} \\
\endfoot
\bottomrule
\endlastfoot
coarse\_perception & 4B & 8B & 1109 & 1.6 & 33 & 15 & 18 \\
logic\_reasoning & 4B & 8B & 412 & 1.2 & 28 & 23 & 5 \\
\end{longtable}
\normalsize
}

\section{Statistical Significance Analysis}
\label{app:significance}

We report one-sided McNemar exact $p$-values as a descriptive measure of paired asymmetry for each of the 375 candidate cases; a case is one subtask--model-pair comparison (a subtask, a smaller model, and a larger model of the same family), and each test is computed over that subtask's shared questions. For each case, we construct a $2 \times 2$ contingency table of per-question outcomes between the smaller and larger model. Under the null hypothesis that both models are equally likely to answer correctly, the number of discordant-inverse outcomes (FlipInv) follows a $\mathrm{Binomial}(\text{FlipInv} + \text{FlipNorm},\, 0.5)$ distribution.

These $p$-values are exploratory, not confirmatory, for two reasons: (1)~cases were pre-selected using the same observed data (Net $\geq 5$, $n \geq 20$), which biases the tested set toward positive effects; and (2)~the tests involve many overlapping model-pair/subtask combinations. They should be read as a descriptive characterization of effect strength, not as family-wise error-controlled evidence. To address (2), we also report Benjamini--Hochberg (BH) adjusted $p$-values (FDR $0.05$), corrected either over the 375 cases or, more conservatively, over the full space of 10{,}434 eligible within-family comparisons ($n \geq 20$) from which the cases were selected.

\begin{table}[h]
\centering
\caption{ McNemar's exact test significance by Net range, uncorrected and after Benjamini--Hochberg (BH) correction at FDR $0.05$. ``BH (cases)'' corrects over the 375 reported cases; ``BH (all)'' corrects over all 10{,}434 eligible comparisons.}
\label{tab:significance}
\begin{tabular}{lccccc}
\toprule
Net Range & Cases & Significant ($p < 0.05$) & \% & BH (cases) & BH (all) \\
\midrule
5--7   & 183 & 19  & 10.4  & 0  & 0 \\
8--10  & 76  & 19  & 25.0  & 3  & 0 \\
11--15 & 68  & 49  & 72.1  & 16 & 0 \\
16--20 & 32  & 30  & 93.8  & 20 & 4 \\
$\geq$21 & 16 & 16  & 100.0 & 15 & 12 \\
\midrule
\textbf{Total} & \textbf{375} & \textbf{133} & \textbf{35.5} & \textbf{54} & \textbf{16} \\
\bottomrule
\end{tabular}
\end{table}

133 of 375 cases (35.5\%) have uncorrected $p < 0.05$, rising monotonically with Net from 10.4\% (Net 5--7) to 100\% (Net $\geq 21$); near the threshold, a net advantage of 5--7 reaches significance only when it arises from few discordant questions (median 9 among the significant cases vs.\ 26 among the rest). After BH correction, 54 cases remain significant over the reported cases and 16 over all eligible comparisons; the latter 16 all have Net $\geq 16$ (\cref{tab:significance}). For scale, the 375 cases are 3.6\% of the 10{,}434 eligible comparisons. The larger model is ahead in 74.8\% of comparisons, and by Net $\leq -5$ in 5{,}000 (47.9\%); under the mirrored one-sided test (the larger model winning the discordant questions), corrected over the same 10{,}434 comparisons, 2{,}478 of these are significant. Scaling therefore helps in most comparisons and detectably reverses in a minority. Under CoT (\cref{app:cot}), 18 of the 375 tests remain BH-significant, all among the 96 persisting cases. Cross-family replication (\cref{sec:cross_family}) provides complementary evidence that these patterns are not confined to a single family.

\section{Cross-Family Replication}
\label{app:cross_family_2}

\cref{tab:cross_family} lists the \ncrossfamily subtask--dataset pairs with inverse scaling in at least three families. \cref{tab:cross_family_2} lists 29 additional subtask--dataset pairs where inverse scaling appears in exactly two of the four families.

\begin{table}[h]
\caption{Subtask--dataset pairs with pairwise inverse scaling in at least three of the four families. ``Max Net'' is the highest Net value across all model pairs and families for that subtask. ``Cases'' is the total number of pairwise cases.}
\label{tab:cross_family}
\centering
\small
\setlength{\tabcolsep}{3pt}
\begin{tabular}{llccc}
\toprule
\textbf{Dataset} & \textbf{Subtask} & \textbf{Families} & \textbf{Max Net} & \textbf{Cases} \\
\midrule
\multirow{5}{*}{BLINK}
& Multi-view Reasoning & 4 & 30 & 22 \\
& Jigsaw & 3 & 36 & 16 \\
& Forensic Detection & 3 & 30 & 13 \\
& Object Localization & 3 & 22 & 10 \\
& Relative Reflectance & 3 & 17 & 11 \\
\midrule
MMBench & Coarse Perception & 3 & 21 & 8 \\
\midrule
\multirow{3}{*}{MMMU test}
& Organic Chemistry & 3 & 15 & 5 \\
& Investment & 3 & 13 & 5 \\
& Thermodynamics & 3 & 9 & 3 \\
\midrule
MMT-Bench & Autonomous Driving & 3 & 23 & 9 \\
MME & Scene & 3 & 16 & 21 \\
MMT-Bench & Relation Reasoning & 3 & 9 & 3 \\
MMT-Bench & Emotion & 3 & 6 & 3 \\
\bottomrule
\end{tabular}
\end{table}

\begin{table}[h]
\caption{Subtask--dataset pairs with pairwise inverse scaling in exactly two families. Max Net is the strongest case across all model pairs in those families.}
\label{tab:cross_family_2}
\centering
\small
\setlength{\tabcolsep}{3pt}
\begin{tabular}{llcc}
\toprule
\textbf{Dataset} & \textbf{Subtask} & \textbf{Families} & \textbf{Max Net} \\
\midrule
MME & celebrity & InternVL, Qwen & 25 \\
MMT-Bench & cross\_image\_matching & InternVL, Qwen & 19 \\
MMBench & attribute\_reasoning & InternVL, Qwen & 18 \\
MMMU test & Electrical Circuit & InternVL, Qwen & 17 \\
BLINK & IQ Test & InternVL, Qwen & 17 \\
MME & artwork & InternVL, Molmo~2 & 17 \\
BLINK & Visual Correspondence & InternVL, Molmo~2 & 17 \\
MMMU test & Materials Science & InternVL, Molmo~2 & 16 \\
MMT-Bench & visual\_recognition & Gemma~3, InternVL & 15 \\
BLINK & Art Style & InternVL, Molmo~2 & 15 \\
MMMU test & Engineering Graphics & InternVL, Qwen & 14 \\
MMMU test & Music & InternVL, Molmo~2 & 14 \\
MMT-Bench & medical\_understanding & Gemma~3, Qwen & 13 \\
MMT-Bench & temporal\_understanding & InternVL, Qwen & 13 \\
MMT-Bench & visual\_illusion & InternVL, Qwen & 12 \\
MMT-Bench & multiple\_image\_analysis & InternVL, Qwen & 9 \\
MMMU test & Artificial Intelligence & InternVL, Qwen & 9 \\
MMMU test & Structural Engineering & InternVL, Qwen & 9 \\
MMT-Bench & threed & Gemma~3, InternVL & 9 \\
BLINK & Semantic Correspondence & InternVL, Molmo~2 & 9 \\
MMMU test & Physiology & Gemma~3, Qwen & 8 \\
MMMU test & Calculus & InternVL, Qwen & 8 \\
MMMU test & Financial Accounting & InternVL, Qwen & 8 \\
MMMU test & Graph Theory & InternVL, Qwen & 7 \\
BLINK & Relative Depth & InternVL, Qwen & 7 \\
AI2D & foodChainsWebs & InternVL, Molmo~2 & 7 \\
MMMU test & Mechanics of Materials & Gemma~3, Molmo~2 & 7 \\
MMT-Bench & action\_recognition & Gemma~3, Qwen & 6 \\
MMMU test & Microbiology & Molmo~2, Qwen & 6 \\
\bottomrule
\end{tabular}
\end{table}

\section{Net Flip Distribution}
\label{app:net_dist}

\cref{tab:net_dist} shows how the number of pairwise cases, unique subtask--dataset pairs, and cross-family replicated subtasks vary with the Net threshold ($n \geq 20$ throughout). The qualitative patterns are stable across thresholds: the same benchmarks and subtasks dominate at all cutoff levels.

\begin{table}[h]
\caption{Robustness to threshold choice: pairwise inverse scaling cases ($n \geq 20$) at different Net thresholds.}
\label{tab:net_dist}
\centering
\small
\begin{tabular}{lcccccc}
\toprule
\textbf{Net threshold} & $\geq 5$ & $\geq 10$ & $\geq 15$ & $\geq 20$ & $\geq 25$ & $\geq 30$ \\
\midrule
Pairwise cases & 375 & 132 & 56 & 21 & 12 & 5 \\
Unique subtask--dataset pairs & 115 & 44 & 25 & 9 & 6 & 3 \\
Cross-family ($\geq$3 families) & 13 & 4 & 2 & 0 & 0 & 0 \\
\bottomrule
\end{tabular}
\end{table}

\section{Evaluation Prompts}
\label{app:prompts}

All models receive identical prompts per dataset. We use \textbf{direct} (zero-shot, no chain-of-thought) prompting throughout to isolate model capacity from reasoning scaffolding.

\paragraph{Multiple-choice datasets.}
For MCQ datasets (MMT-Bench, MMMU, ScienceQA, MMBench), the prompt is constructed as:

\begin{quote}
\small
\texttt{[Question]}\\
\texttt{A. [Option A]}\\
\texttt{B. [Option B]}\\
\texttt{C. [Option C]}\\
\texttt{...}\\
\texttt{Please select the correct answer from the options above. Respond with only the letter of the correct option. Do not explain. Answer:}
\end{quote}

\noindent MMBench additionally prepends a \texttt{hint} field (image context description) before the question when available. MMMU questions may contain embedded \texttt{<image N>} placeholders referencing multiple sub-images within the question text.

\paragraph{Open-ended questions.}
MMMU contains a subset of fill-in-the-blank questions without multiple-choice options. For these, the suffix is replaced with:

\begin{quote}
\small
\texttt{Answer concisely with just the final value or term. Answer:}
\end{quote}

\paragraph{Yes/No dataset (MME).}
MME questions are presented as-is without any additional suffix. Each question expects a ``Yes'' or ``No'' response.

\paragraph{Pre-formatted datasets.}
AI2D, MathVista, and BLINK use pre-formatted prompts from their respective source distributions. AI2D uses the VLMEvalKit-processed version with masked diagrams. MathVista uses the \texttt{query} field from AI4Math/MathVista, which includes task-specific hints and answer format instructions. BLINK prompts are used verbatim from the original benchmark.

\subsection{Error Taxonomy Judge}

For the error analysis (\cref{sec:taxonomy}), we classify each inverse scaling flip question using GPT-5.4 and Claude Opus~4.6 as multimodal annotators. GPT-5.4 is accessed via the OpenAI API (model identifier: \texttt{gpt-5.4}). Claude Opus~4.6 is accessed via the Anthropic API (model identifier: \texttt{claude-opus-4-6}). Each question is sent with its associated image(s) and the following text prompt (temperature = 0, max tokens = 200). The prompt is blinded to model identities; models are labeled ``Model A'' (correct) and ``Model B'' (incorrect) without revealing their sizes:

\begin{quote}
\small
\begin{verbatim}
You are analyzing VLM (Vision-Language Model)
benchmark results. Two models answered this
question. Model A answered correctly and Model B
answered incorrectly. Your job is to figure out
why Model B failed.

Question: {question}
Options: {options}
Correct Answer: {answer}

Model A response: {response_small}
  -> CORRECT

Model B response: {response_large}
  -> INCORRECT

Categorize Model B's failure into exactly
ONE of these categories:

1. GENUINE_ERROR - Model B made a clear mistake --
   whether through visual misidentification, wrong
   reasoning, incorrect knowledge, falling for a
   plausible distractor, or simply picking the wrong
   answer. This covers any case where the model
   genuinely got it wrong.
2. FORMAT_MISMATCH - Model B effectively gives the
   correct answer content, but in a form that would
   not match the benchmark answer format (e.g.,
   numeric value instead of option letter, or full
   option text instead of just the letter). If the
   response is self-contradictory or clearly commits
   to the wrong answer, use GENUINE_ERROR.
3. BAD_ANNOTATION_OR_AMBIGUOUS - Use this only when
   there is strong evidence that the ground-truth
   answer is wrong, the question is genuinely
   ambiguous, or multiple answers would reasonably
   be acceptable. Do NOT use this category just
   because Model B's answer seems plausible; if the
   benchmark answer is still clearly better
   supported, use GENUINE_ERROR.

Reply with ONLY a JSON object (no markdown,
no extra text):
{"category": "CATEGORY_NAME",
 "confidence": "high/medium/low",
 "explanation": "One concise sentence
  explaining why."}
\end{verbatim}
\end{quote}

\noindent Each annotator assigns exactly one category per question. We report GPT-5.4 results in the main text and use Claude Opus~4.6 as a secondary check. The two judges agree on 90.8\% of questions (Cohen's $\kappa = 0.30$; \cref{sec:taxonomy}). We treat these annotations as exploratory qualitative labels, not as validated ground truth.

\section{Human Annotation Validation}
\label{app:human_annotation}

Three authors independently annotated a stratified subsample of 127 taxonomy questions using the same three categories (Genuine Error, Bad Annotation / Ambiguous, Format Mismatch). The sample overrepresents questions flagged as Bad Annotation by either LLM judge to ensure sufficient coverage of the minority class.

\begin{table}[h]
\caption{Human annotation results on 127 sampled taxonomy questions.}
\label{tab:human_annotation}
\centering
\small
\begin{tabular}{lccc}
\toprule
& \textbf{Genuine Error} & \textbf{Bad Annot.} & \textbf{Format Mis.} \\
\midrule
Annotator 1 & 105 (82.7\%) & 22 (17.3\%) & 0 \\
Annotator 2 & 110 (86.6\%) & 17 (13.4\%) & 0 \\
Annotator 3 & 101 (79.5\%) & 26 (20.5\%) & 0 \\
\midrule
Majority vote & 107 (84.3\%) & 20 (15.7\%) & 0 \\
\bottomrule
\end{tabular}
\end{table}

\paragraph{Inter-annotator agreement.} Pairwise agreement ranged from 86.6\% to 92.1\% ($\kappa = 0.53$--$0.74$). Fleiss $\kappa$ across all three raters was 0.65 (substantial agreement).

\paragraph{Human vs.\ LLM judges.} The human majority vote agreed with GPT-5.4 on 88.2\% and with Claude Opus~4.6 on 84.3\% of questions. Humans labeled 15.7\% of questions as Bad Annotation or Ambiguous, compared to 2.6\% (GPT) and 10.5\% (Claude), indicating that human annotators apply a more liberal standard for annotation quality issues than either LLM judge. Nonetheless, all five raters (three human, two LLM) agree that the large majority of sampled flips reflect genuine model errors.
\section{Restricting to Comparisons of at Least Twofold Size}
\label{app:size_ratio}

The main analysis compares every pair of sizes within a family. Some of these pairs are close in size: consecutive dense checkpoints such as InternVL 8B and 14B differ by less than a factor of two, and when models are ordered by total parameter count, a Mixture-of-Experts checkpoint can rank above a dense one while activating fewer parameters per token (Qwen 30B-A3B activates 3B, the dense 8B activates 8B). Here we repeat the case count under two restrictions that remove such comparisons: (i) the larger model has at least twice the \emph{total} parameters of the smaller, which removes 3 pairs (InternVL 30B-A3B$\to$38B, InternVL 8B$\to$14B, Qwen3-VL 30B-A3B$\to$32B); and (ii) the larger model has at least twice the \emph{active} parameters, which additionally removes the 9 pairs in which the larger model is an MoE with fewer than twice the active parameters of the smaller (InternVL 14B$\to$30B-A3B, InternVL 2B$\to$30B-A3B, InternVL 38B$\to$241B-A28B, InternVL 4B$\to$30B-A3B, InternVL 8B$\to$30B-A3B, Qwen3-VL 2B$\to$30B-A3B, Qwen3-VL 32B$\to$235B-A22B, Qwen3-VL 4B$\to$30B-A3B, Qwen3-VL 8B$\to$30B-A3B). \cref{tab:ratio_total,tab:ratio_active} give the resulting counts in the format of \cref{tab:pairwise_summary}; \cref{tab:ratio_summary} summarizes them next to the unrestricted analysis.

\begin{table}[h]
\caption{Pairwise inverse scaling cases over eligible comparisons, restricted to pairs in which the larger model has at least twice the total parameters of the smaller. Gemma~3 and Molmo~2 have no pair below this ratio, so their counts are unaffected.}
\label{tab:ratio_total}
\centering
\small
\setlength{\tabcolsep}{4pt}
\begin{tabular}{lrrrrr}
\toprule
\textbf{Dataset} & \textbf{Qwen3-VL} & \textbf{InternVL} & \textbf{Gemma~3} & \textbf{Molmo~2} & \textbf{Total} \\
\midrule
MMT-Bench & 29/448 & 46/832 & 6/96 & 1/32 & 82/1408 \\
MMMU (dev\_val+test) & 24/1512 & 50/2808 & 8/324 & 6/108 & 88/4752 \\
ScienceQA & 4/420 & 0/780 & 10/90 & 0/30 & 14/1320 \\
MME & 10/196 & 18/364 & 0/42 & 2/14 & 30/616 \\
AI2D & 0/182 & 7/338 & 0/39 & 1/13 & 8/572 \\
MathVista & 1/70 & 1/130 & 0/15 & 0/5 & 2/220 \\
BLINK & 36/196 & 44/364 & 2/42 & 8/14 & 90/616 \\
MMBench & 1/84 & 9/156 & 0/18 & 2/6 & 12/264 \\
\midrule
\textbf{Total} & \textbf{105/3108} & \textbf{175/5772} & \textbf{26/666} & \textbf{20/222} & \textbf{326/9768} \\
\bottomrule
\end{tabular}
\end{table}

\begin{table}[h]
\caption{As \cref{tab:ratio_total}, restricted to pairs in which the larger model has at least twice the active parameters of the smaller. Gemma~3 and Molmo~2 are dense and have no pair below this ratio, so their counts are unaffected.}
\label{tab:ratio_active}
\centering
\small
\setlength{\tabcolsep}{4pt}
\begin{tabular}{lrrrrr}
\toprule
\textbf{Dataset} & \textbf{Qwen3-VL} & \textbf{InternVL} & \textbf{Gemma~3} & \textbf{Molmo~2} & \textbf{Total} \\
\midrule
MMT-Bench & 20/352 & 24/704 & 6/96 & 1/32 & 51/1184 \\
MMMU (dev\_val+test) & 26/1188 & 33/2376 & 8/324 & 6/108 & 73/3996 \\
ScienceQA & 7/330 & 0/660 & 10/90 & 0/30 & 17/1110 \\
MME & 8/154 & 16/308 & 0/42 & 2/14 & 26/518 \\
AI2D & 0/143 & 3/286 & 0/39 & 1/13 & 4/481 \\
MathVista & 0/55 & 0/110 & 0/15 & 0/5 & 0/185 \\
BLINK & 21/154 & 27/308 & 2/42 & 8/14 & 58/518 \\
MMBench & 1/66 & 7/132 & 0/18 & 2/6 & 10/222 \\
\midrule
\textbf{Total} & \textbf{83/2442} & \textbf{110/4884} & \textbf{26/666} & \textbf{20/222} & \textbf{239/8214} \\
\bottomrule
\end{tabular}
\end{table}

\begin{table}[h]
\caption{ Summary of the two restrictions, with the unrestricted analysis for reference. ``Larger ahead'' counts comparisons with Net $\leq -5$; ``Replicated'' is the number of subtask--dataset pairs with a case in at least three families.}
\label{tab:ratio_summary}
\centering
\small
\begin{tabular}{lrrrrr}
\toprule
\textbf{Pairs kept} & \textbf{Comparisons} & \textbf{Cases} & \textbf{Rate} & \textbf{Larger ahead} & \textbf{Replicated} \\
\midrule
All pairs & 10434 & 375 & 3.6\% & 5000 (47.9\%) & 13 \\
Total parameters $\geq 2\times$ & 9768 & 326 & 3.3\% & 4856 (49.7\%) & 11 \\
Active parameters $\geq 2\times$ & 8214 & 239 & 2.9\% & 4190 (51.0\%) & 11 \\
\bottomrule
\end{tabular}
\end{table}

The total-parameter restriction changes little: 326 of the 375 cases remain, at a rate of 3.3\%, and 11 of the 13 replicated subtasks survive. The active-parameter restriction removes the pairs in which the case rate is highest: the 2,220 excluded comparisons contain 136 cases (6.1\%) and no replicated subtask. Under this restriction 239 cases remain from 8,214 comparisons (2.9\%), the larger model is ahead by Net $\leq -5$ in 51.0\% of comparisons, and 11 of the 13 replicated subtasks survive; BLINK falls from 95 to 58 cases and MathVista to none. Cross-family replication is therefore insensitive to how the pairs are restricted, whereas the case count depends on whether a checkpoint that activates fewer parameters is counted as the larger model.

\section{Option-Order Sensitivity}
\label{app:option_order}

All main-text results keep each benchmark's provided option order, the standard protocol, which preserves comparability with published numbers. Within a pair, both checkpoints receive the same prompt, option order, decoding, and answer-extraction pipeline, so a presentation effect shared by both models cannot by itself produce Net~$>0$; a differential sensitivity of the two checkpoints to option position could. To check this, we re-evaluate the six BLINK cases among the ten strongest cases of \cref{tab:top_cases} under deterministic reorderings of the answer options: for two-option subtasks the single reversal, and for four-option subtasks the reversal and the three cyclic shifts. The reordering is applied identically to every question of the subtask and to both models of the pair, the gold letter is remapped accordingly, and every other setting is unchanged. The identity ordering is re-run as well, so that the effect of reordering can be read against the run-to-run variation of greedy decoding on different hardware (the re-runs use a different cluster from the main evaluation).

\begin{table}[h]
\caption{ Option-order sensitivity for the six BLINK cases of \cref{tab:top_cases}. Each row is one deterministic reordering of the answer options; ``identity'' replays the original order. Acc is accuracy (\%) of the smaller (S) and larger (L) model; FI, FN and Net as in the main text.}
\label{tab:option_order}
\centering
\small
\setlength{\tabcolsep}{3pt}
\begin{tabular}{cllllrrrrr}
\toprule
\textbf{\#} & \textbf{Family} & \textbf{Subtask} & \textbf{Pair} & \textbf{Ordering} & \textbf{Acc S} & \textbf{Acc L} & \textbf{FI} & \textbf{FN} & \textbf{Net} \\
\midrule
\multirow{2}{*}{1} & \multirow{2}{*}{Qwen3-VL} & \multirow{2}{*}{Jigsaw} & \multirow{2}{*}{4B vs.\ 30B-A3B} & identity & 80.7 & 64.0 & 38 & 13 & $+25$ \\
 & & &  & reversal & 76.0 & 59.3 & 38 & 13 & $+25$ \\
\midrule
\multirow{2}{*}{2} & \multirow{2}{*}{Qwen3-VL} & \multirow{2}{*}{Jigsaw} & \multirow{2}{*}{4B vs.\ 32B} & identity & 80.7 & 60.7 & 40 & 10 & $+30$ \\
 & & &  & reversal & 76.0 & 70.0 & 29 & 20 & $+9$ \\
\midrule
\multirow{2}{*}{3} & \multirow{2}{*}{Molmo~2} & \multirow{2}{*}{Jigsaw} & \multirow{2}{*}{4B vs.\ 8B} & identity & 80.0 & 58.7 & 44 & 12 & $+32$ \\
 & & &  & reversal & 80.7 & 54.0 & 49 & 9 & $+40$ \\
\midrule
\multirow{4}{*}{4} & \multirow{4}{*}{InternVL3.5} & \multirow{4}{*}{Forensic Detection} & \multirow{4}{*}{2B vs.\ 30B-A3B} & identity & 23.5 & 2.3 & 28 & 0 & $+28$ \\
 & & &  & reversal & 20.5 & 6.8 & 22 & 4 & $+18$ \\
 & & &  & shift $+1$ & 22.0 & 3.8 & 26 & 2 & $+24$ \\
 & & &  & shift $+2$ & 22.7 & 7.6 & 28 & 8 & $+20$ \\
\midrule
\multirow{2}{*}{5} & \multirow{2}{*}{Qwen3-VL} & \multirow{2}{*}{Multi-view Reasoning} & \multirow{2}{*}{8B vs.\ 32B} & identity & 51.9 & 24.1 & 44 & 7 & $+37$ \\
 & & &  & reversal & 49.6 & 28.6 & 40 & 12 & $+28$ \\
\midrule
\multirow{2}{*}{8} & \multirow{2}{*}{Qwen3-VL} & \multirow{2}{*}{Multi-view Reasoning} & \multirow{2}{*}{2B vs.\ 32B} & identity & 51.9 & 24.1 & 43 & 6 & $+37$ \\
 & & &  & reversal & 46.6 & 28.6 & 38 & 14 & $+24$ \\
\bottomrule
\end{tabular}
\end{table}

The smaller model's net advantage stays positive in all 17 runs, 11 reorderings and 6 identity replays. Reordering changes the magnitude, most visibly for Qwen 4B vs.\ 32B on Jigsaw, where the reversal reduces Net from $+30$ to $+9$ because the 32B improves by 9\pp under the reversed order, and for InternVL on Forensic Detection, where the 30B-A3B gains a few correct answers under every reordering but never more than 8\pp. The identity replays also differ from the main-evaluation values in \cref{tab:top_cases} by up to 11 Net points in either direction, which bounds the run-to-run variation of greedy decoding across hardware; the reordering effects are of the same order. Option position therefore modulates the size of these cases but does not reverse them.

\section{Vision Encoders and Image Preprocessing}
\label{app:encoders}

Within-family comparisons hold the training recipe roughly fixed but not necessarily the vision encoder. \cref{tab:encoders} lists, for each model, the vision encoder as stated in the family's technical report or model card and confirmed against the released configuration files. Two families change encoder across sizes: Qwen3-VL uses SigLIP2-Large (300M) for the 2B and 4B models and SigLIP2-SO-400M for the larger ones~\citep{qwen3vl}, and InternVL3.5 uses InternViT-300M up to the 14B model and 30B-A3B and InternViT-6B for the 38B and 241B-A28B~\citep{internvl35}; in both, the encoder is trained jointly, so same-tier weights may still differ across sizes. Gemma~3 shares one frozen 417M SigLIP encoder across all three sizes~\citep{gemma3}, and Molmo~2 uses the same SigLIP~2 SO-400M configuration for both sizes~\citep{molmo2}. All models use their native dynamic-resolution or tiling preprocessing at the defaults of their released processors, with no image resizing on our side.

\begin{table}[h]
\caption{ Vision encoders by model. ``Same weights'' indicates whether the encoder weights are shared across the sizes in the row according to the family's report.}
\label{tab:encoders}
\centering
\small
\setlength{\tabcolsep}{4pt}
\begin{tabular}{llllc}
\toprule
\textbf{Family} & \textbf{Models} & \textbf{Encoder} & \textbf{Params} & \textbf{Same weights} \\
\midrule
Qwen3-VL & 2B, 4B & SigLIP2-Large & 0.3B & no (jointly trained) \\
Qwen3-VL & 8B, 30B-A3B, 32B, 235B-A22B & SigLIP2-SO-400M & 0.4B & no (jointly trained) \\
InternVL3.5 & 1B, 2B, 4B, 8B, 14B, 30B-A3B & InternViT-300M & 0.3B & no (jointly trained) \\
InternVL3.5 & 38B, 241B-A28B & InternViT-6B & 5.5B & no (jointly trained) \\
Gemma~3 & 4B, 12B, 27B & SigLIP & 0.4B & yes (frozen) \\
Molmo~2 & 4B, 8B & SigLIP~2 SO-400M & 0.4B & same configuration \\
\bottomrule
\end{tabular}
\end{table}

Of the 375 cases, 271 (72.3\%) compare two models with the same encoder tier (Qwen3-VL 68, InternVL3.5 157, Gemma~3 26, Molmo~2 20); the 46 Gemma~3 and Molmo~2 cases compare models whose encoders are shared or configuration-identical, so only the language model differs in size. The remaining 104 cases cross an encoder tier (Qwen3-VL 54, InternVL3.5 50), and in every InternVL3.5 instance the larger model is the one with the roughly twenty times larger encoder. Restricting the cross-family analysis to same-tier cases leaves 10 of the 13 replicated subtasks, including BLINK Multi-view Reasoning in all four families.

\section{Chain-of-Thought Control Experiment}
\label{app:cot}

\paragraph{Design.} We re-evaluate both models of every case under CoT prompting on all shared questions of the subtask: 375 cases, 66{,}698 matched question instances, and 92{,}694 model--question evaluations over 522 (model, dataset, subtask) runs. The CoT prompt appends a step-by-step instruction to the dataset prompt; for multiple-choice questions it reads: ``Answer the preceding multiple choice question. The last line of your response should be of the following format: `Answer: \$LETTER' (without quotes) where LETTER is one of options. Think step by step before answering.'' Decoding is greedy with a 4{,}096-token budget, and answers are scored with the same pipeline as the direct condition. Because both models are re-run, Net is recomputed from the CoT outputs rather than adjusted.

\paragraph{Result.} \cref{tab:cot_family} reports the outcome per family and \cref{tab:cot_dataset} per dataset. CoT attenuates inverse scaling but does not remove it. The median Net falls from $+8$ to $0$; 96 of 375 cases (25.6\%) retain Net~$\geq 5$, 178 fall below the threshold, and 101 reverse sign (Net $\leq -5$). The larger model recovers 3{,}640 of the 8{,}361 questions (43.5\%) it had lost under direct prompting. At the same time, the smaller model's accuracy decreases under CoT in 251 of 375 cases (mean $-3.9$\pp, vs.\ $+5.2$\pp for the larger model), so part of the reduction in Net reflects the smaller model losing its advantage rather than the larger model gaining.

\paragraph{Where it persists.} Persistence differs markedly across benchmarks. On MME and BLINK, close to half of the cases survive CoT (48.6\% and 42.1\%), whereas on MMMU the effect largely disappears: 7.6\% of test-split cases persist and 41 of 105 reverse. One reading is that CoT helps more where the difficulty lies in reasoning over recognized content than where it lies in extracting the answer from the image, but we have not tested this directly.

\paragraph{Significance.} Applying the test of \cref{app:significance} to the CoT outcomes, 44 of the 96 persisting cases have uncorrected $p < 0.05$, and 18 remain significant after BH correction over the 375 CoT tests (14 with Net $\geq 21$), vs.\ 54 of the same cases under direct prompting.

\begin{table}[h]
\caption{ Both-model chain-of-thought control, by family. \textbf{Persist} counts cases still meeting Net~$\geq 5$ under CoT; \textbf{reverse} counts cases where the larger model gains a net advantage of 5 or more. \textbf{Recovery} is the share of the questions the larger model lost in direct mode that it answers correctly under CoT.}
\label{tab:cot_family}
\centering
\small
\begin{tabular}{lrrrrrrr}
\toprule
\textbf{Family} & \textbf{Cases} & \multicolumn{2}{c}{\textbf{Median Net}} & \textbf{Persist} & \textbf{Dissolve} & \textbf{Reverse} & \textbf{Recovery} \\
\cmidrule(lr){3-4}
 & & direct & CoT & & & & \\
\midrule
Qwen3-VL      & 122 & $+8$  & $0$  & 35 (28.7\%) & 59  & 28  & 40.1\% \\
InternVL3.5   & 207 & $+7$  & $0$  & 54 (26.1\%) & 99  & 54  & 45.4\% \\
Gemma~3       & 26  & $+7$  & $-2$ & 2 (7.7\%)   & 14  & 10  & 46.9\% \\
Molmo~2       & 20  & $+10$ & $-4$ & 5 (25.0\%)  & 6   & 9   & 39.7\% \\
\midrule
\textbf{All}  & \textbf{375} & $\mathbf{+8}$ & $\mathbf{0}$ & \textbf{96 (25.6\%)} & \textbf{178} & \textbf{101} & \textbf{43.5\%} \\
\bottomrule
\end{tabular}
\end{table}

\begin{table}[h]
\caption{ Both-model chain-of-thought control, by dataset. Inverse scaling survives CoT on perception-centric benchmarks and collapses on knowledge-centric ones.}
\label{tab:cot_dataset}
\centering
\small
\begin{tabular}{lrrrrrr}
\toprule
\textbf{Dataset} & \textbf{Cases} & \multicolumn{2}{c}{\textbf{Median Net}} & \textbf{Persist} & \textbf{Reverse} & \textbf{Recovery} \\
\cmidrule(lr){3-4}
 & & direct & CoT & & & \\
\midrule
MME          & 35  & $+7$  & $+4$  & 17 (48.6\%) & 7  & 46.2\% \\
BLINK        & 95  & $+10$ & $+3$  & 40 (42.1\%) & 17 & 46.4\% \\
AI2D$^{*}$   & 9   & $+6$  & $+1$  & 3 (33.3\%)  & 1  & 14.8\% \\
MMT-Bench    & 87  & $+7$  & $0$   & 22 (25.3\%) & 16 & 37.3\% \\
ScienceQA    & 17  & $+8$  & $-2$  & 2 (11.8\%)  & 6  & 59.0\% \\
MMMU dev\_val & 10  & $+6$  & $-1$  & 1 (10.0\%)  & 2  & 48.9\% \\
MMMU test & 105 & $+7$  & $-3$  & 8 (7.6\%)   & 41 & 42.6\% \\
MMBench      & 14  & $+12$ & $-14$ & 1 (7.1\%)   & 10 & 49.4\% \\
MathVista    & 3   & $+7$  & $+5$  & 2 (66.7\%)  & 1  & 47.1\% \\
\bottomrule
\end{tabular}
\\[2pt]
\footnotesize $^{*}$AI2D cases are dominated by InternVL, which answers with a bare letter on 98--99\% of AI2D questions under the CoT prompt (the dataset prompt itself asks for the letter), so these are effectively direct-mode results.
\end{table}

\section{Qualitative Examples of Inverse Scaling}
\label{app:examples}

We present representative examples where a smaller model answers correctly but a larger model from the same family fails. All models were prompted with direct answering (``Respond with only the letter''), yet several produce chain-of-thought reasoning unprompted (Examples 1, 2, 5, 6, and 9) and still arrive at the wrong answer, suggesting the errors are not due to insufficient reasoning. Long responses are shortened, with omissions marked ``\ldots''; Examples 9 and 12 also show answers from the chain-of-thought control (\cref{app:cot}), and full responses are provided in the supplementary material.

\paragraph{Example 1: BLINK / Jigsaw (Molmo 4B vs 8B, Net = 31).}

\begin{center}
\includegraphics[width=0.3\textwidth]{figures/examples/ex2_jigsaw_1.jpg}
\includegraphics[width=0.3\textwidth]{figures/examples/ex2_jigsaw_2.jpg}
\includegraphics[width=0.3\textwidth]{figures/examples/ex2_jigsaw_3.jpg}
\end{center}

\begin{description}[leftmargin=1em,labelwidth=0em]
\item[\textbf{Question:}] Which image is the missing part in the first image?
\item[\textbf{Answer:}] B
\item[\textbf{Molmo 4B} {\color{green!60!black}\ding{51}}:] ``Let's carefully analyze this question step by step: 1.~The first image has a missing lower right corner, creating a black square. 2.~We need to compare this with the second and third images to determine which would fit best\ldots'' $\rightarrow$~\textbf{B}
\item[\textbf{Molmo 8B} {\color{red}\ding{55}}:] ``Before answering, let's consider the details of each image\ldots Image~1 shows two men in an outdoor setting, with the lower right corner missing\ldots Image~3 shows a group of Asian individuals\ldots'' $\rightarrow$~\textbf{C}
\end{description}
\noindent\textbf{Analysis:} Both models reason step by step, but the larger model selects the wrong piece.

\paragraph{Example 2: BLINK / Multi-view Reasoning (Qwen 2B vs 32B, Net = 29).}

\begin{center}
\includegraphics[width=0.45\textwidth]{figures/examples/ex3_multiview_1.jpg}
\includegraphics[width=0.45\textwidth]{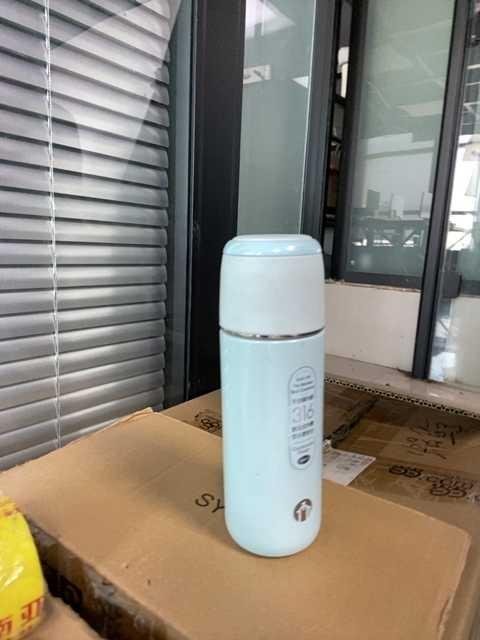}
\end{center}

\begin{description}[leftmargin=1em,labelwidth=0em]
\item[\textbf{Question:}] The images are frames from a video. Is the camera moving left or right?
\item[\textbf{Answer:}] B (right)
\item[\textbf{Qwen 2B} {\color{green!60!black}\ding{51}}:] ``To determine the direction of the camera movement, we need to analyze the change in the object's position\ldots The light blue thermos is the central subject. In the first image, it is positioned more towards the center. In the second image, it has been shifted slightly to the right\ldots'' $\rightarrow$~\textbf{B}
\item[\textbf{Qwen 32B} {\color{red}\ding{55}}:] ``To determine the direction of the camera movement, we compare the two images. In the first image, the object (a light blue thermos) is positioned slightly to the right\ldots The vertical blind on the left is now more [visible]\ldots'' $\rightarrow$~\textbf{A}
\end{description}
\noindent\textbf{Analysis:} Both models analyze the same visual cues but the larger model infers the opposite direction.

\paragraph{Example 3: MME / Celebrity (InternVL 8B vs 14B, Net = 25).}

\begin{center}
\includegraphics[width=0.4\textwidth]{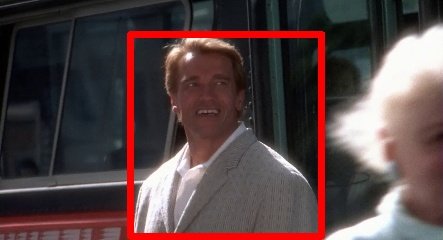}
\end{center}

\begin{description}[leftmargin=1em,labelwidth=0em]
\item[\textbf{Question:}] Is the actor inside the red bounding box called Arnold Schwarzenegger?
\item[\textbf{Answer:}] Yes
\item[\textbf{InternVL 8B} {\color{green!60!black}\ding{51}}:] Yes
\item[\textbf{InternVL 14B} {\color{red}\ding{55}}:] No
\end{description}
\noindent\textbf{Analysis:} A straightforward visual identification failure: the person is clearly Arnold Schwarzenegger.

\paragraph{Example 4: MMT-Bench / Autonomous Driving (Qwen 2B vs 235B-A22B, Net = 19).}

\begin{center}
\includegraphics[width=0.5\textwidth]{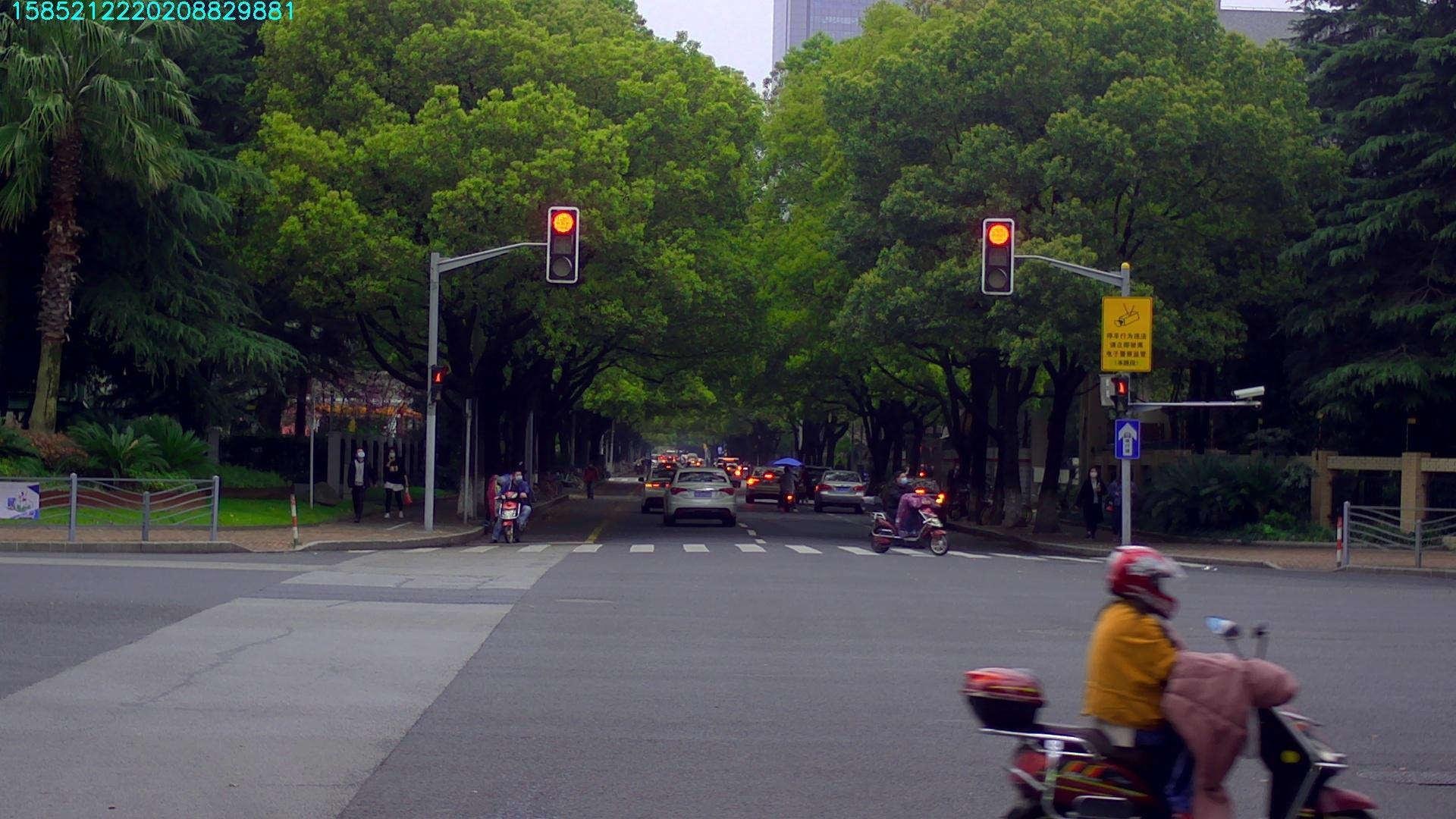}
\end{center}

\begin{description}[leftmargin=1em,labelwidth=0em]
\item[\textbf{Question:}] Which traffic lights can you see from the image?
\item[\textbf{Options:}]~\\[2pt]
\hspace*{1em}A.~Wait on \\[2pt]
\hspace*{1em}\textbf{B.~Red~\ding{51}} \\[2pt]
\hspace*{1em}C.~Yellow \\[2pt]
\hspace*{1em}D.~Green
\item[\textbf{Qwen 2B} {\color{green!60!black}\ding{51}}:] B
\item[\textbf{Qwen 235B-A22B} {\color{red}\ding{55}}:] C (Yellow)
\end{description}
\noindent\textbf{Analysis:} The smallest Qwen model correctly identifies red traffic lights. The largest (235B MoE) says yellow, a safety-critical error in an autonomous driving context.

\paragraph{Example 5: BLINK / Visual Correspondence (Molmo 4B vs 8B, Net = 16).}

\begin{center}
\includegraphics[width=0.45\textwidth]{figures/examples/ex9_blink_corresp_1.jpg}
\includegraphics[width=0.45\textwidth]{figures/examples/ex9_blink_corresp_2.jpg}
\end{center}

\begin{description}[leftmargin=1em,labelwidth=0em]
\item[\textbf{Question:}] Which point is corresponding to the reference point?
\item[\textbf{Answer:}] A
\item[\textbf{Molmo 4B} {\color{green!60!black}\ding{51}}:] A
\item[\textbf{Molmo 8B} {\color{red}\ding{55}}:] ``The point in the first image that corresponds to the point in the second image is Point~B. In the first image, there's a red circle with a black dot in the center, labeled `REF'\ldots'' $\rightarrow$~\textbf{B}
\end{description}
\noindent\textbf{Analysis:} The larger model reasons about the spatial correspondence but arrives at the wrong point.

\paragraph{Example 6: MathVista / Geometry (InternVL 8B vs 14B, Net = 9).}

\begin{center}
\includegraphics[width=0.35\textwidth]{figures/examples/ex10_mathvista_geometry.jpg}
\end{center}

\begin{description}[leftmargin=1em,labelwidth=0em]
\item[\textbf{Question:}] AB is the diameter of $\odot$O, point D is on the extended line of AB, passing point D is the tangent of $\odot$O, and the tangent point is C. If $\angle A = 25^\circ$, then $\angle D = $ ?
\item[\textbf{Options:}]~\\[2pt]
\hspace*{1em}A.~$25^\circ$ \\[2pt]
\hspace*{1em}\textbf{B.~$40^\circ$~\ding{51}} \\[2pt]
\hspace*{1em}C.~$50^\circ$ \\[2pt]
\hspace*{1em}D.~$65^\circ$
\item[\textbf{InternVL 8B} {\color{green!60!black}\ding{51}}:] ``Connect OC. Since CD is tangent to the circle, OC $\perp$ CD, so $\angle$OCD $= 90^\circ$. Since OA $=$ OC, $\angle$A $= \angle$OCA $= 25^\circ$. Thus $\angle$DOC $= 50^\circ$, and $\angle$D $= 90^\circ - 50^\circ = 40^\circ$.'' $\rightarrow$~\textbf{B}
\item[\textbf{InternVL 14B} {\color{red}\ding{55}}:] ``Since AB is the diameter, $\angle$ACB $= 90^\circ$. Given $\angle$A $= 25^\circ$, $\angle$CBA $= 90^\circ - 25^\circ = 65^\circ$. Since CD is a tangent to the circle\ldots'' $\rightarrow$~wrong answer
\end{description}
\noindent\textbf{Analysis:} Both models attempt multi-step geometric reasoning. The smaller model correctly uses tangent-radius perpendicularity and isosceles triangle properties, while the larger model misapplies the tangent-chord relationship.

\noindent\begin{minipage}{\linewidth}
\paragraph{Example 7: MMT-Bench / Action Recognition (Gemma~3 4B vs 12B, Net = 6).}

\begin{center}
\includegraphics[width=0.28\textwidth]{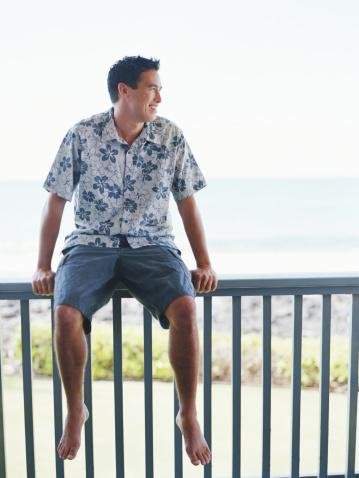}
\end{center}
\end{minipage}

\begin{description}[leftmargin=1em,labelwidth=0em]
\item[\textbf{Question:}] What is the action type shown in this image? (A) lie (B) squat (C) run (D) sit
\item[\textbf{Answer:}] D (sit)
\item[\textbf{Gemma~3 4B} {\color{green!60!black}\ding{51}}:] D
\item[\textbf{Gemma~3 12B} {\color{red}\ding{55}}:] B
\end{description}
\noindent\textbf{Analysis:} The man is plainly seated on the railing with his legs hanging down; the larger model labels the pose as squatting.

\noindent\begin{minipage}{\linewidth}
\paragraph{Example 8: ScienceQA / Astronomy (Qwen 30B-A3B vs 32B, Net = 5).}

\begin{center}
\includegraphics[width=0.4\textwidth]{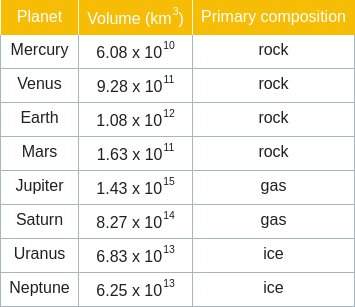}
\end{center}
\end{minipage}

\begin{description}[leftmargin=1em,labelwidth=0em]
\item[\textbf{Question:}] Is the following statement about our solar system true or false? Earth's volume is more than ten times as great as Mars's volume. (A) false (B) true
\item[\textbf{Answer:}] A (false)
\item[\textbf{Qwen 30B-A3B} {\color{green!60!black}\ding{51}}:] A
\item[\textbf{Qwen 32B} {\color{red}\ding{55}}:] B
\end{description}
\noindent\textbf{Analysis:} Ten times the volume of Mars is $1.63 \times 10^{12}$\,km$^3$, more than the volume of Earth ($1.08 \times 10^{12}$\,km$^3$), so the statement is false. The answer requires reading two values from the table and one comparison.

\noindent\begin{minipage}{\linewidth}
\paragraph{Example 9: BLINK / Forensic Detection (Qwen 4B and 8B vs 235B-A22B, Net = 20 and 14; under CoT, 16 and 17).}

\begin{center}
\includegraphics[width=0.22\textwidth]{figures/examples/ex13_blink_forensic_1.jpg}
\includegraphics[width=0.22\textwidth]{figures/examples/ex13_blink_forensic_2.jpg}
\includegraphics[width=0.22\textwidth]{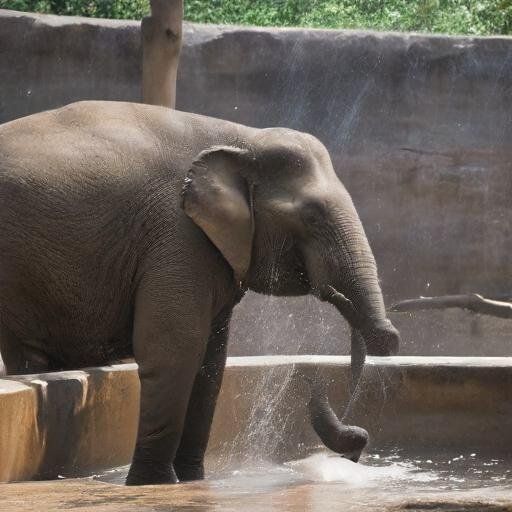}
\includegraphics[width=0.22\textwidth]{figures/examples/ex13_blink_forensic_4.jpg}
\end{center}
\end{minipage}

\begin{description}[leftmargin=1em,labelwidth=0em]
\item[\textbf{Question:}] Which image is most likely to be a real photograph? (A) the first image (B) the second image (C) the third image (D) the fourth image
\item[\textbf{Answer:}] A (the first image)
\item[\textbf{Qwen 4B} {\color{green!60!black}\ding{51}}:] ``\ldots{} The other images, especially B, C, and D, have elements that suggest digital manipulation or AI generation --- the overly clean water sprays, the unnatural lighting, and the overly perfect textures. Therefore, the most likely to be a real photograph is: (A) the first image.'' \quad (CoT: \textbf{A})
\item[\textbf{Qwen 8B} {\color{green!60!black}\ding{51}}:] ``\ldots{} The other three images exhibit telltale signs of AI generation or heavy digital manipulation, such as unrealistic water effects, inconsistent lighting, and unnatural textures. Therefore, the correct answer is (A) the first image.'' \quad (CoT: \textbf{A})
\item[\textbf{Qwen 235B-A22B, CoT} {\color{red}\ding{55}}:] ``Let's analyze each image step by step for signs of realism \ldots{} Image 1: \ldots{} The pillars have very uniform, repetitive texture --- almost too perfect \ldots{} Suspicious for AI generation due to texture uniformity and lack of natural interaction with environment. \ldots{} Image 3: \ldots{} Water spray is dynamic and realistic \ldots{} Conclusion: Image 3 stands out as the most realistic \ldots{} Final Answer: (C) the third image.'' \quad (direct: \textbf{C})
\end{description}
\noindent\textbf{Analysis:} Both smaller models identify the real photograph with and without chain-of-thought. The largest Qwen model reasons through every image and rejects the real one as ``suspicious for AI generation'', under both direct and CoT prompting.

\noindent\begin{minipage}{\linewidth}
\paragraph{Example 10: MMMU test / Investment (Gemma~3 4B vs 12B, Net = 13).}

\begin{center}
\includegraphics[width=0.6\textwidth]{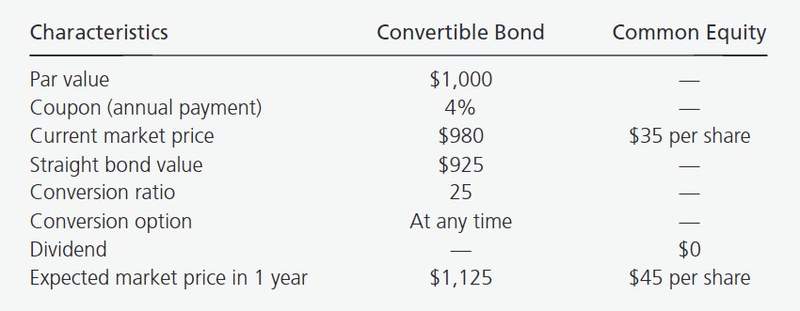}
\end{center}
\end{minipage}

\begin{description}[leftmargin=1em,labelwidth=0em]
\item[\textbf{Question:}] Rich McDonald, CFA, is evaluating his investment alternatives in Ytel Incorporated by analyzing a Ytel convertible bond and Ytel common equity. Characteristics of the two securities are given in the exhibit. Calculate, based on the exhibit, the expected 1-year rate of return for the Ytel common equity. (A) 27.39\% (B) 26.44\% (C) 28.57\%
\item[\textbf{Answer:}] C (28.57\%)
\item[\textbf{Gemma~3 4B} {\color{green!60!black}\ding{51}}:] C
\item[\textbf{Gemma~3 12B} {\color{red}\ding{55}}:] A
\end{description}
\noindent\textbf{Analysis:} With no dividend, the expected return is the price change, $(\$45 - \$35)/\$35 = 28.57\%$. The smaller model selects this value; the larger model selects a distractor.

\noindent\begin{minipage}{\linewidth}
\paragraph{Example 11: MME / Landmark (InternVL 1B vs 2B, Net = 29).}

\begin{center}
\includegraphics[width=0.28\textwidth]{figures/examples/ex15_mme_landmark.jpg}
\end{center}
\end{minipage}

\begin{description}[leftmargin=1em,labelwidth=0em]
\item[\textbf{Question:}] Is this a picture of Church of Our Lady of Sorrows in Riga? Please answer yes or no.
\item[\textbf{Answer:}] Yes
\item[\textbf{InternVL 1B} {\color{green!60!black}\ding{51}}:] Yes.
\item[\textbf{InternVL 2B} {\color{red}\ding{55}}:] ``No, this is not the Church of Our Lady of Sorrows in Riga. The building in the image appears to be a church with a distinctive architectural style, but it does not match the specific characteristics of the Church of Our Lady of Sorrows in Riga\ldots''
\end{description}
\noindent\textbf{Analysis:} The larger model rejects the correct landmark and justifies the rejection at length. This pair is among the ten strongest cases (\cref{tab:top_cases}).

\noindent\begin{minipage}{\linewidth}
\paragraph{Example 12: BLINK / Relative Reflectance (Qwen 32B vs 235B-A22B, Net = 17; under CoT, 8).}

\begin{center}
\includegraphics[width=0.65\textwidth]{figures/examples/ex16_blink_reflectance.jpg}
\end{center}
\end{minipage}

\begin{description}[leftmargin=1em,labelwidth=0em]
\item[\textbf{Question:}] Consider the surface color (color without shading) of the two points in the image. Which one is darker, or the color is about the same? (A) A is darker (B) B is darker (C) About the same
\item[\textbf{Answer:}] B (B is darker)
\item[\textbf{Qwen 32B, CoT} {\color{green!60!black}\ding{51}}:] ``Point A is located on the wall, just above the sofa, between the two framed pictures. The wall is painted a light beige or cream color \ldots{} Point B is located on the dark gray or charcoal-colored sofa \ldots{} Therefore, Point B has a darker surface color than Point A. Final Answer: (B) B is darker.'' \quad (direct: \textbf{B})
\item[\textbf{Qwen 235B-A22B, CoT} {\color{red}\ding{55}}:] ``Point A is located on the backrest of the dark gray couch. Point B is located on the seat cushion of the same couch. Both points are on the same piece of furniture \ldots{} the surface color (albedo) should be the same. Therefore, the correct answer is: (C) About the same.'' \quad (direct: \textbf{C})
\end{description}
\noindent\textbf{Analysis:} The largest model misplaces point A on the sofa rather than the wall and reasons correctly from the wrong premise, under both direct and CoT prompting. The 32B locates both points correctly.

\end{document}

%% file: math_commands.tex
\usepackage{amsmath,amsfonts,bm}

\def\eqref#1{equation~\ref{#1}}

\def\1{\bm{1}}

\DeclareMathAlphabet{\mathsfit}{\encodingdefault}{\sfdefault}{m}{sl}
\SetMathAlphabet{\mathsfit}{bold}{\encodingdefault}{\sfdefault}{bx}{n}

